\documentclass{article}
\usepackage{iclr2027_conference,times}
\newcommand{\buildmode}{preprint} 
\newcommand{\buildsubmission}{submission}
\newcommand{\buildpreprint}{preprint}
\ifx\buildmode\buildsubmission\else\iclrfinalcopy\fi

\usepackage{latexsym}
\usepackage[T1]{fontenc}
\usepackage[utf8]{inputenc}
\usepackage{microtype}
\usepackage{graphicx}
\usepackage{wrapfig}
\usepackage{float}
\usepackage[font=small,labelfont=bf]{caption}
\usepackage{xcolor}
\definecolor{sig}{HTML}{095792}\definecolor{sigdk}{HTML}{0A3F6E}\definecolor{sigmid}{HTML}{81A3CF}\definecolor{siglt}{HTML}{CCDFF7}
\definecolor{fail}{HTML}{C85138}\definecolor{faillt}{HTML}{FDDFD7}
\definecolor{mdl}{HTML}{6B7076}\definecolor{mdllt}{HTML}{EEEFF1}\definecolor{ink}{HTML}{222222}
\definecolor{sigrow}{HTML}{E6EFFA} 
\usepackage{hyperref}
\hypersetup{colorlinks=true,linkcolor=sigdk,citecolor=sigdk,urlcolor=sigdk} 
\usepackage{etoc} 
\usepackage{placeins} 
\usepackage{url}
\usepackage{booktabs}
\usepackage{makecell} 
\usepackage{colortbl}
\usepackage{enumitem}
\usepackage{amsmath}
\usepackage{amssymb}
\usepackage{pifont}
\newcommand{\cmark}{\ding{51}} 
\newcommand{\xmark}{\ding{55}}
\usepackage{algorithm}
\usepackage{algpseudocode}
\usepackage{tikz}
\usepackage[most]{tcolorbox}
\newtcolorbox{claimbox}{enhanced,colback=sigrow,colframe=sigdk,boxrule=0.6pt,arc=2pt,left=6pt,right=6pt,top=4pt,bottom=4pt,
  title={In one sentence},fonttitle=\bfseries\small,coltitle=white,colbacktitle=sigdk,attach boxed title to top left={xshift=6pt,yshift=-2pt},
  boxed title style={arc=1.5pt,boxrule=0pt},before skip=6pt,after skip=6pt}
\usetikzlibrary{arrows.meta,positioning,decorations.pathmorphing,fit,backgrounds,calc,patterns}
\usepackage[textsize=footnotesize]{todonotes}

\DeclareRobustCommand{\src}[1]{{\scriptsize\textcolor{mdl}{\nolinkurl{#1}}}}         



\title{A Tilted Bowl Is Not a Slippery Slope:\\ Compressing Looped Models}

\author{Steven Kolawole$^{1,2}$ \quad Pearse Jim$^{2}$ \quad Opegbemi M. Busoye$^{2}$ \quad Glory Bagai$^{2}$ \quad Virginia Smith$^{1}$ \\[4pt]
\normalfont $^1$Carnegie Mellon University \quad $^2$ML Collective \\[2pt]
\normalfont Correspondence: \texttt{skolawol@cs.cmu.edu}
}

\begin{document}
\maketitle
\ifx\buildmode\buildpreprint\lhead{Preprint. Under review.}\fi
\etocdepthtag.toc{maintag}

\begin{abstract}
Looped models reason by applying the same block of weights many times, so compressing that block saves memory traffic
on every loop. Compressed looped models, however, often collapse, and the collapse is usually blamed on rounding error
that accumulates from loop to loop. In this work we test that account on more than 30 models from five families and find, to our surprise, that it
holds only for loops that never settle. When a loop settles, a fixed rounding error does not accumulate. It moves the
point where the loop settles, much as tilting a bowl moves where a ball comes to rest, and the answer is lost only when
the shift is larger than the readout tolerates. This picture lets us predict which models fail from a single label-free
measurement, and it tells us why failed models recover: their loops still settle, so a few final loops with 8-bit
weights bring the answer back. Motivated by these findings, we build a controller that stops when the model's halting head fires and then finishes
with 8-bit loops. On Sudoku-Extreme and Maze-Hard it beats fixed-depth inference by up to 15 points under a third of
the weight traffic.
\end{abstract}
\begin{figure}[h!]
\centering
\includegraphics[width=\linewidth]{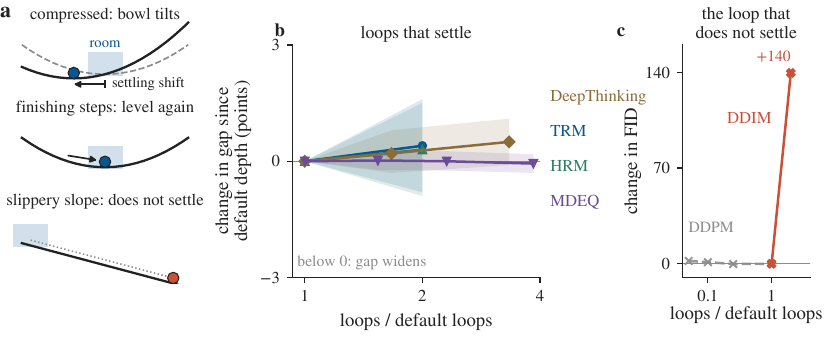}   
\caption{\textbf{Thinking longer does not hurt a compressed model whose loop settles; a loop that does not settle
piles up error.} (a)~Compression tilts the bowl, so the loop settles slightly off target. The answer survives while
the settling shift stays within the room, and finishing steps level the bowl again. On a slippery slope, every loop
carries the ball further.
(b)~Change in the gap between a 4-bit model and its own full precision as the loops go on past the default depth,
one model per family; a value below zero means the gap widens. No interval lies below zero. (c)~DDIM, the one loop
that does not settle, adds 140 FID over the same range. Paired 95\% intervals; the looped language models are in Table~\ref{tab:lms}; protocol in
App.~\ref{app:depth}.}
\label{fig:teaser}
\end{figure}
\section{Introduction}
\label{sec:intro}

Looped models get their depth from repetition. Instead of stacking new layers, they apply one block of weights again
and again, from Universal Transformers \citep{dehghani2019universal} and looped transformers \citep{saunshi2025latent}
to recurrent-depth language models \citep{geiping2025recurrentdepth,zhu2025ouro} and small recursive reasoners
\citep{wang2025hrm,jolicoeur2025trm}. Repetition keeps them small, and TRM solves hard Sudoku and maze puzzles with
about 7 million parameters. It moves their cost into memory traffic instead, because every loop reads the same weights
again. Compressing the weights therefore saves on every loop, which makes quantization the natural lever for running a
looped model cheaply \citep{nagel2021whitepaper,frantar2023gptq,lin2024awq}.

Quantized looped models, however, fail abruptly. The MLP-mixer TRMs we train solve about 81\% of Sudoku-Extreme
puzzles with 4-bit weights and one scale per output channel, yet under 8\% when the same weights share one scale per
tensor (\S\ref{sec:room}). The usual explanation is that the rounding error enters the block on every loop and
accumulates as the model thinks. LoopQ builds a loop-aware quantizer on this account \citep{fang2026loopq}, and
concurrent work traces the collapse of TRM-style reasoners to a bias that accumulates with reuse
\citep{ingolfsson2026quantizing}. If error accumulated, every extra loop would add risk, and a deployer would have to
trade precision against thinking time. \emph{Does the error really accumulate as a compressed model thinks?}

To find out, we run compressed models far past their default depth and track their gap to full precision loop by
loop, on more than 30 models from five families (Table~\ref{tab:families}). The answer surprised us. A compressed
model that works at its default depth keeps working as it thinks longer (Figure~\ref{fig:teaser}b), and the result
reaches public looped language models, where Huginn with 3-bit weights answers 3\% of GSM8K problems against 42\% at
full precision and sixteen precise final loops bring it back to 40\% (\S\ref{sec:lms}). Error accumulates only in
deterministic DDIM sampling, whose loop never settles (Figure~\ref{fig:teaser}c). The usual explanation holds only
for loops that never settle.

Think of a looped model as a ball rolling into a bowl. Each loop rolls the ball further down, and the spot where it
comes to rest is the answer. Rounding the weights tilts the bowl, so the
ball comes to rest at a slightly different spot, a move we call the \emph{settling shift}. Rounding is a \emph{fixed
error}, the same on every loop, so the tilt is fixed and further loops carry the ball no further, the classical fact
that a fixed perturbation moves the fixed point of a contraction only a bounded distance \citep{banach1922operations}.
A \emph{fresh error}, drawn anew on every loop, nudges the ball in a new direction each time, and further loops let it
settle back; only on a slippery slope, where nothing pulls the ball back, does every loop carry it further.

The same picture lets us predict which models fail. Around the old resting spot lies a region of states that read out
the same answer, which we call the model's \emph{room}, and the \emph{push} is the error that one compressed loop makes
at the settled state. An answer is lost when the push exceeds the room, much as a classifier keeps its prediction while
a perturbation stays within its margin over its sensitivity \citep{tsuzuku2018lipschitz}. With one room shared by all
models and a label-free measurement of each model's sensitivity, we predict which models collapse before deploying
them (\S\ref{sec:room}).

It also explains why a failed model can be recovered. A collapsed model has settled in the wrong place, but it still
settles. A few final loops with more precise weights, which we call \emph{finishing steps}, pull its state back, as
mixed-precision iterative refinement solves cheaply and then refines precisely
\citep{carson2018threeprecisions,higham2022mixed}. Eight-bit weights are enough for these steps, and a law we stated
before the runs predicts how much they recover (\S\ref{sec:return}).
\begin{claimbox}
\centering\emph{Compressed looped models do not drift as they think; they settle slightly off target,\\
and a few precise loops bring them back.}
\end{claimbox}

From these findings we build a controller that stops a puzzle when the model's halting head fires and then takes a
few 8-bit finishing steps. On Sudoku-Extreme and Maze-Hard it beats fixed-depth inference by up to 15 points under a
third of the weight traffic, without a full-precision copy of the compressed weights (\S\ref{sec:controller}). We
summarize our
contributions below.
\begin{enumerate}[leftmargin=1.4em,itemsep=1pt,topsep=2pt]
\item \textbf{Compressed loops settle.} On 33 models from five families, a compressed model that works at its default
  depth keeps working as it thinks longer, and the size of the error decides whether it collapses
  (\S\ref{sec:depth}, \S\ref{sec:fixedfresh}).
\item \textbf{Failure is predictable.} A model fails when its push exceeds its room, and one room shared by the
  puzzle, convnet and implicit families turns a label-free sensitivity into each model's tolerance (\S\ref{sec:room}).
\item \textbf{Failure is recoverable.} A collapsed model still settles, and a few 8-bit finishing steps bring it back,
  in looped language models as well (\S\ref{sec:return}, \S\ref{sec:lms}).
\item \textbf{A controller.} Stopping at the halting head and finishing with 8-bit loops beats fixed depth on
  Sudoku-Extreme and Maze-Hard (\S\ref{sec:controller}).
\end{enumerate}
\section{Setup}
\label{sec:setup}

\subsection{Looped models and settling}
\label{sec:setup-settle}

A looped model applies one weight-tied map $f_\theta$ to a latent state again and again. The input $x$ enters at every
loop, and a readout $g_\theta$ turns the last state into the answer:
\begin{equation}
  z_{t+1} \;=\; f_\theta(z_t,\, x), \qquad \hat y \;=\; g_\theta(z_T).
  \label{eq:loop}
\end{equation}
The loop \emph{settles} when its late steps shrink, $\lVert z_{t+1}-z_t\rVert < \lVert z_t-z_{t-1}\rVert$, and the
state it approaches is the \emph{settled state} $z^\star(\theta,x)$. A contraction always settles, and a fixed
perturbation moves its settled state by a bounded amount \citep{banach1922operations}. Learned loops need not contract,
however. Some of our Sudoku TRMs expand certain directions even at full precision and still settle
(Appendix~\ref{app:return}). We measure settling on every compressed model instead of assuming it.

\subsection{Push, sensitivity and room}
\label{sec:setup-push}

Compression replaces the weights $\theta$ by $\tilde\theta$, and the \emph{settling shift} is the resulting move of
the settled state. Its source is the \emph{push}, the error that one compressed loop makes at the settled state:
\begin{equation}
  \delta \;=\; \frac{\lVert f_{\tilde\theta}(z^\star, x) - f_\theta(z^\star, x)\rVert}{\lVert z^\star\rVert}
  \;\approx\; S\,\sigma .
  \label{eq:push}
\end{equation}
Here $\sigma$ is the weight error relative to the scale of each tensor, and $S$ is the model's \emph{sensitivity}.
We measure $S$ without labels, by adding relative weight noise at a small level to the loop, running one loop from the
settled state on 256 unlabelled inputs, and taking the median push per unit of noise over three draws; $S\sigma$
tracks the push measured on the compressed models (Spearman 0.91, Appendix~\ref{app:widthlaw}). For a contraction with rate $L<1$
the settling shift is at most $\delta/(1-L)$. The \emph{room} $\delta^\star$ is the largest push that leaves the
answer unchanged, so a model tolerates weight error up to $\sigma_{1/2}=\delta^\star/S$, the level at which its
accuracy halves. We estimate a single room shared by all models (\S\ref{sec:room}).

\subsection{Compression, fixed error and fresh error}
\label{sec:setup-compress}

We quantize weights after training by round-to-nearest, with one scale per tensor (t), per output channel (c) or per
group of 32 input weights (g32); w4t is 4-bit per-tensor, w3g32 3-bit in groups of 32 (Appendix~\ref{app:protocol-quant}).
Quantization is the compression these models survive, since pruning a quarter of TRM's structure, or distilling it
into a smaller student, leaves it unable to solve a single puzzle (Appendix~\ref{app:other-compression}). Rounding is a \emph{fixed error}, since the same rounded weights serve every loop. To separate the size of an error
from its fixedness, we replace rounding by Gaussian noise of relative size $\sigma$ in two ways: added to the weights
once and kept, a fixed error, or added to the input of every linear layer and drawn anew at every call, a \emph{fresh
error}. A third variant gives the noise a consistent sign, the bias that concurrent work holds responsible for the
collapse \citep{ingolfsson2026quantizing}.

\subsection{Models and measurements}
\label{sec:setup-protocol}

\begin{table}[t]
\caption{\textbf{Five families, 33 models, one repeated map each.} The family is the unit of the claims; the boundary
case, diffusion, is the one loop that need not settle. Every model, its default depth and its metric are in
Appendix~\ref{app:models}.}
\label{tab:families}
\centering\scriptsize
\setlength{\tabcolsep}{4pt}
\begin{tabular}{@{}p{0.17\textwidth}p{0.40\textwidth}p{0.15\textwidth}p{0.22\textwidth}@{}}
\toprule
Family (models) & Models & Task & What loops \\
\midrule
Puzzle reasoners (24) & TRM with an MLP or an attention mixer, HRM \citep{jolicoeur2025trm,wang2025hrm}; 19 TRMs we train at several widths and seeds & Sudoku-Extreme, Maze-Hard & one block, $H\times L$ cycles per supervision step \\
Recurrent convnets (1) & DeepThinking \citep{bansal2022overthinking} & mazes of size 13 and 33 & a recall block, input re-injected \\
Implicit models (3) & MDEQ-Small, MDEQ-XL \citep{bai2020mdeq}; DEQ-Transformer \citep{bai2019deq} & ImageNet, WikiText-103 & a solver, run to a fixed point \\
Looped language models (4) & Huginn-3.5B \citep{geiping2025recurrentdepth}; Ouro-1.4B, Ouro-2.6B \citep{zhu2025ouro}; Recurrent-OLMo-2 \citep{mcleish2025retrofitted} & GSM8K & a core block, 4--32 recurrences; full precision is bf16 \\
Diffusion, the boundary (1) & DDPM UNet, sampled by DDPM \citep{ho2020ddpm} and DDIM \citep{song2021ddim} & CIFAR-10 & the denoiser; DDPM draws fresh noise each step, DDIM is deterministic \\
\bottomrule
\end{tabular}
\end{table}

Table~\ref{tab:families} lists the five families. Each has its own depth dial, and we call all of them loops.

If error accumulated, the gap between a compressed model and full precision would widen as the loops go on. We
compare every compressed model with full precision on the same examples at the same number of loops, and
call a gap \emph{widening} only if the paired bootstrap interval of its change lies below zero
\citep{efron1993bootstrap}. A compressed model \emph{survives} if it keeps at least half of full-precision accuracy at
the default depth, and \emph{collapses} otherwise. \emph{Fidelity} $\varphi$, the cosine between the compressed and
full-precision final states, gives a label-free view of the same comparison. We ran on two GPU platforms with different
software stacks; each exhibit comes from one platform, and no number pools them (Appendix~\ref{app:protocol-stats}).
\section{The settling shift}
\label{sec:shift}

In the picture of the introduction, compression tilts the bowl once, and the loop still settles, slightly off target.
We test the picture in five steps: whether error accumulates with the loops (\S\ref{sec:depth}), what decides collapse
(\S\ref{sec:fixedfresh}), whether a collapsed model comes back (\S\ref{sec:return}), which models collapse
(\S\ref{sec:room}), and whether looped language models behave alike (\S\ref{sec:lms}). One compressed model runs
through the section: HRM on Sudoku-Extreme with 2-bit weights in groups of 32. It solves 14\% of the puzzles, where its
full-precision version solves 48\%.

\subsection{A compressed model that settles keeps its accuracy as it thinks longer}
\label{sec:depth}

\paragraph{Thinking longer does not widen the gap.}
A puzzle reasoner has three depth dials, the inner cycles $L$, the outer cycles $H$ and the supervision steps that
repeat them (Table~\ref{tab:families}). We run each compressed puzzle reasoner far past its default depth on each
dial and track its gap to full precision. In
Figure~\ref{fig:teaser}b the gap of one model per family holds past the default depth. Across every series, none of 22
widens along the outer cycles, 21 of 22 hold or narrow along the supervision steps, and a second platform with new
evaluation rows agrees, with none of 23 widening (Appendix~\ref{app:depth-series}).
The exceptions are the surviving format closest to collapse, per-tensor 4-bit weights, which loses a few points along
supervision steps in some runs, and the fast inner cycles, where some series widen and we make no claim.

\paragraph{A collapsed implicit model passes the standard health check.}
A deep equilibrium solver declares success when its residual is small. The collapsed DEQ-Transformer reaches a residual
481 times smaller than its full-precision version, so the check passes a model that has failed. Its solver has
converged, only to the wrong state. Compressed DeepThinking solvers, MDEQ and the surviving DEQ-Transformer lose nothing
to extra iterations (Figure~\ref{fig:teaser}b; Appendix~\ref{app:depth-deq}).

\paragraph{Error does accumulate where the loop never settles.}
At 4-bit grouped weights, the FID of deterministic DDIM samples to full-precision samples rises from 40 at 10 steps to
304 at 100, whereas ancestral DDPM sampling, which draws fresh noise at every step, stays flat
(Figure~\ref{fig:teaser}c; Appendix~\ref{app:depth-diffusion}). The usual account of accumulating error is right for
a loop that never settles, and only there. Ouro is the looped language model of this kind, since it loses accuracy with extra
loops even at full precision (\S\ref{sec:lms}).

\paragraph{The concurrent Maze result is a clipping choice.}
Concurrent work reports that 4-bit per-tensor weights and activations collapse Sudoku but not Maze, and that Maze
improves with depth \citep{ingolfsson2026quantizing}. The clipping rule for activations decides Maze. A scale set by
the largest calibration value gives 0.0\% accuracy, and a 99.99th-percentile scale gives 82.2\%. The concurrent work
does not state its statistic, and under the percentile rule the Maze gap narrows over the first loops and then holds,
the shape of a settling shift (Appendix~\ref{app:depth-ingolfsson}).

For a deployer, a compressed model that works at its default depth can think as long as its budget allows.

\subsection{Error size sets the cliff; fixedness decides repair}
\label{sec:fixedfresh}

\begin{figure}[t]
\centering
\includegraphics[width=\linewidth]{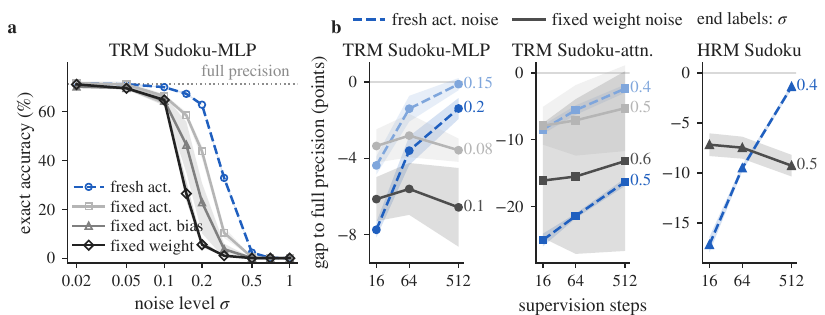}
\caption{\textbf{The size of the error sets the cliff; whether the error stays fixed decides whether more loops
repair it.} (a)~Accuracy against noise level for TRM Sudoku-MLP. Zero-mean fresh noise, which has nothing to
accumulate, produces the same cliff as fixed noise, only later. (b)~Gap to full precision as the supervision steps
grow, for fresh error, drawn anew at every call, and fixed error, added to the weights once, matched on fidelity; the
number at each line end is the noise level. Fresh error heals with more loops in every draw; fixed error stays, except
in some attention draws. Means and 95\% intervals over noise draws; protocol in App.~\ref{app:fixedfresh}.}
\label{fig:fixedfresh}
\end{figure}

\paragraph{Noise with nothing to accumulate still makes the cliff.}
Zero-mean noise drawn anew at every call has no bias to build up. Injected into TRM Sudoku-MLP, it still takes the
model from 62.9\% to 2.3\% as its size grows from 0.2 to 0.5 of each tensor's scale (Figure~\ref{fig:fixedfresh}a), and
it does the same on every puzzle reasoner. A fixed draw or a consistent sign moves the cliff to smaller noise but does
not create it (Appendix~\ref{app:fixedfresh-arms}). Collapse needs no accumulating bias; an error of sufficient size is
enough.

\paragraph{More loops repair a fresh error and leave a fixed one.}
Size decides collapse, but not whether more loops help. We match fixed and fresh error on fidelity and run each model
far past its default depth. Fresh error closes part of its gap in all 30 draws at the smaller noise levels; fixed
error closes it in none of 15 draws on the MLP mixer and none of 15 on HRM (Figure~\ref{fig:fixedfresh}b). Rounding
behaves like fixed error, since no quantized MLP or HRM model closes its gap with more loops
(Appendix~\ref{app:fixedfresh-noise}).

\paragraph{The attention mixer repairs some fixed errors.}
On the attention mixer, 5 of 15 fixed-error draws close part of their gap, and the surviving 3-bit formats close it
fully. Attention maps that sharpen to one position at the settled state would explain this; in a probe we specified in
advance, they stay far from one-hot and the attention error does not shrink (Appendix~\ref{app:fixedfresh-onehot}).

A deployer cannot loop a rounding error away; its size must stay small.

\subsection{A collapsed model still settles, and a few precise loops bring it back}
\label{sec:return}

\paragraph{A collapsed model still settles.}
If compression broke the loop, a collapsed model's late steps would stop shrinking. In 88\% of collapsed models they
still shrink, at nearly the rate of surviving ones; our running example's late steps are each 0.93 times the one
before (Appendix~\ref{app:return_contraction}). The late ratio is the contraction rate $L$ of
\S\ref{sec:setup-settle}, measured along the trajectory. Compression moves the settled state and leaves the loop settling. Concurrent
work infers settling from drift \citep{ingolfsson2026quantizing}; we measure it, on the models that fail.

\paragraph{The failed state is still within reach.}
A collapsed model might have settled so far away that the full-precision loop cannot bring it back. We start the
full-precision loop from the compressed settled state and from points several settling shifts beyond it. In 29 of 64
collapsed models the full-precision loop recovers the answer even from the farthest point we try
(Appendix~\ref{app:return_prereg}). Collapse happens with the state still in reach.

\begin{figure}[t]
\centering
\includegraphics[width=\linewidth]{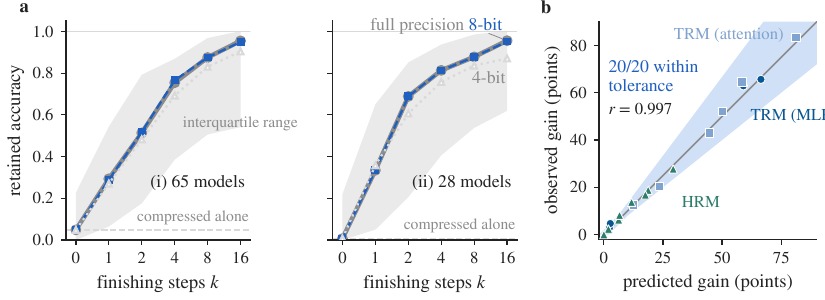} 
\caption{\textbf{A collapsed model comes back, 8-bit finishing is enough, and the gain is predicted in advance.}
(a)~Share of full-precision accuracy that collapsed models retain after $k$ finishing steps with full-precision,
8-bit or 4-bit weights; $k{=}0$ is the compressed model alone. The 8-bit curve lies on the full-precision curve.
Medians over models with the interquartile band; (i) and (ii) are the two platforms. (b)~Gain predicted by the
finishing law against the observed gain, one point per model; all 20 fall inside the tolerance set before the run
(band). Protocol in App.~\ref{app:return}.}
\label{fig:return}
\end{figure}

\paragraph{A few 8-bit loops bring the answer back.}
We hand the settled state of each collapsed model to a precise loop for a few finishing steps. The median collapsed
model keeps 5\% of its full-precision accuracy on its own and 96\% after 16 finishing steps, on both platforms
(Figure~\ref{fig:return}a). Finishing with 8-bit weights recovers as much, 95\%. The running example climbs from 14\% to
45\% with 16 steps at 8 bits, against 48\% at full precision. We had predicted that four steps would do; the median
collapsed model needs eight to sixteen, as a loop whose late steps shrink by 0.93 removes only a fraction of the shift
per step (Appendix~\ref{app:return_k}).

\paragraph{We predicted the gain before the runs.}
An example gains from finishing exactly when compression lost it and its state returns under the full-precision loop,
so the expected gain is the share of such examples. We wrote this law down before the finishing runs, with the other
predictions of this section; Appendix~\ref{app:return_prereg} lists each with its outcome, the three that failed
included. All 20 models on which the law can be scored fall within the tolerance we set, and predicted and observed
gains correlate at 0.997 (Figure~\ref{fig:return}b; Appendix~\ref{app:return_law}).

A failed compressed model needs no retraining; a few 8-bit loops at the end recover it.

\subsection{A model fails when its push exceeds its room}
\label{sec:room}

\begin{figure}[t]
\centering
\includegraphics[width=\linewidth]{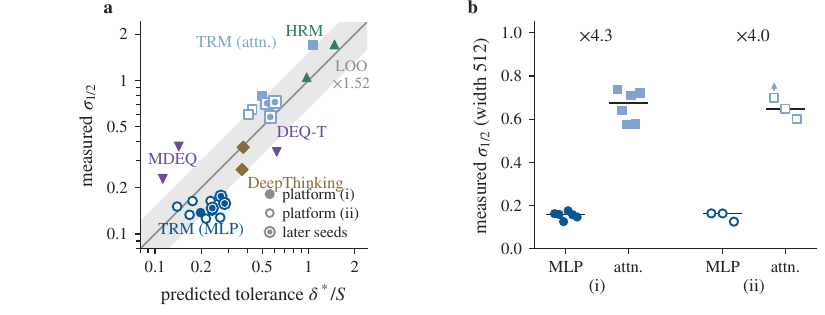} 
\caption{\textbf{Who fails is predictable before deployment: a model tolerates weight error up to its room divided by
its sensitivity.} (a)~Tolerance predicted from label-free sensitivity and one room shared by all models, against the
measured weight error at which accuracy halves; one point per model and task (18 models on 19 pairs, $R^2{=}0.72$ in
log scale). Filled and open markers are the two platforms, ringed markers the seeds held out of the fit; the band is
the leave-one-out median error. (b)~Measured tolerance of width-512 TRMs, one point per seed. The attention mixer
tolerates four times the error of the MLP mixer on both platforms. Fits in App.~\ref{app:widthlaw}.}
\label{fig:widthlaw}
\end{figure}

\paragraph{One number predicts which models fail.}
If a model fails when its push exceeds its room, it tolerates weight error up to $\delta^\star/S$
(\S\ref{sec:setup-push}). We measure each model's sensitivity $S$ without labels and fit one room shared by all models.
That single number predicts the tolerance of 18 models from three families, with $R^2=0.72$ on a log scale and a
leave-one-out error of a factor of 1.5 at the median, and a free fit recovers the predicted slope
(Figure~\ref{fig:widthlaw}a). Six mixer seeds that played no part in the fit land inside the law's stated error band
(Appendix~\ref{app:widthlaw_seeds}). The room is fitted, not measured; the obvious measured candidate, the distance at
which answers change along the compression direction, does not predict tolerance, because the settling shift grows
faster than linearly with the error (Appendix~\ref{app:widthlaw_negative}).

\paragraph{The mixer sets the room.}
Attention-mixer TRMs tolerate four times the weight error of MLP-mixer TRMs, on both platforms
(Figure~\ref{fig:widthlaw}b). For an MLP-mixer model the quantizer alone decides the outcome. Our MLP-mixer TRMs solve
about 81\% of puzzles with one scale per output channel at 4 bits and under 8\% with one scale per tensor. The mixer
gap is also the law's weak point, since a single room overestimates the tolerance of MLP-mixer models and
underestimates that of attention (Appendix~\ref{app:widthlaw_smix}).

\paragraph{Training lowers the push.}
Quantization-aware training with learned step sizes \citep{esser2020lsq} widens the tolerance of the full-precision model itself, and it does so as the law allows, by cutting the
sensitivity by about 1.7 times and leaving the room unchanged (Appendix~\ref{app:widthlaw_lsq}).

\paragraph{A cheap check flags failures.}
Fidelity orders compressed models by collapse within every model, and a cut set on 50 labelled examples places a new
model's boundary (Appendix~\ref{app:widthlaw_detector}). For each input, the halting head is a safe signal, since an answer on which it fires in a
surviving model is almost always correct.

A deployer can rank candidate models and formats before running any labelled evaluation.

\subsection{Looped language models}
\label{sec:lms}

\paragraph{Huginn and Recurrent-OLMo-2 settle and come back.}
We quantize the looped block of four public looped language models (Table~\ref{tab:lms}). Huginn with 3-bit weights
answers 3\% of GSM8K problems, against 42\% at full precision; with the second half of its loops at full precision, it
answers 40\%. Recurrent-OLMo-2 goes from 2\% to 45\%, against 41\% at full precision. In both models the states of
surviving compressed models return to the full-precision answer, and accuracy does not fall with more loops
(Appendix~\ref{app:lms_return}).

\begin{table}[t]
\caption{\textbf{Looped language models that settle come back when their last loops run at full precision; Ouro,
which does not settle, marks the boundary.} GSM8K accuracy (\%) at full precision (bf16), for the compressed model
alone, and with the second half of its loops at full precision (+precise). One platform, $n{=}100$ per entry (Wilson
half-width about 10 points); settling tests, fidelity and prompts in App.~\ref{app:lms}.}
\label{tab:lms}
\centering\small
\setlength{\tabcolsep}{3.5pt}
\begin{tabular}{@{}lccccccc@{}}
\toprule
 & & & \multicolumn{2}{c}{w4c} & \multicolumn{2}{c}{w3a} \\
\cmidrule(lr){4-5}\cmidrule(lr){6-7}
Model (loops) & Settles & bf16 & alone & +precise & alone & +precise \\
\midrule
Huginn-3.5B (32)       & \cmark & 42 & 33 & 37 & 3 & 40 \\ 
Recurrent-OLMo-2 (32)  & \cmark & 41 & 35 & 46 & 2 & 45 \\ 
Ouro-1.4B (4)          & \xmark & 73 & 7  & 66 & 0 & 0  \\ 
Ouro-2.6B (4)          & \xmark & 81 & 78 & 74 & 0 & 50 \\ 
\bottomrule
\end{tabular}
\end{table}
\paragraph{Ouro does not settle.}
Ouro is trained for four loops, and even at full precision its accuracy falls when it loops longer. Precise final loops
repair some compressed Ouro models, but the picture gives no guarantee there, and 8-bit final loops on a surviving
compressed Ouro gain nothing over the compressed model alone at the same weight traffic (Appendix~\ref{app:lms_settle}).

\paragraph{The fidelity check needs a cut per model.}
Fidelity orders collapse within each language model, but the cut differs between models, and no label-free quantity
we tried predicts it. The divergence of next-token distributions from full precision separates collapse in every
model (Appendix~\ref{app:lms_detect}). Teacher-forced loss alone is not
enough, since a small rise in loss coexists with collapsed generation (Appendix~\ref{app:lms_detect}).

\paragraph{In summary.}
A compressed looped model that settles fails in one way, by settling off target. Thinking longer costs it nothing, the
size of the error decides whether it collapses, a few 8-bit loops bring a collapsed model back, and its push against its
room predicts in advance whether it collapses at all.
\section{A controller built from the settling shift}
\label{sec:controller}

The picture suggests two levers. Past the room a trajectory can reach the right answer and then settle off target, so
stopping at the right moment keeps the answer; and a model that has settled off target can be brought back by
finishing steps. A controller built from these two levers is also a test of the picture.

\subsection{Stop, then finish}
\label{sec:controller-levers}

\paragraph{The halting head says when to stop.}
TRM and HRM carry a \emph{halting head}, a small readout of the latent state trained to predict whether the current
answer is already correct \citep{jolicoeur2025trm,wang2025hrm}. The controller stops a puzzle at the first loop at
which the head fires. The firing rule is set once per model
on full-precision development rows and kept for every compressed model, so the controller never sees compressed
labels, and the head is compressed with the other weights (Appendix~\ref{app:controller-head}). Stopping early also
avoids overthinking, the loss of an answer the trajectory has already passed \citep{bansal2022overthinking}.

\paragraph{Finishing steps bring the state back, at 8 bits.}
After it stops, or at a depth cap, the controller takes a few finishing steps with an 8-bit copy of the weights, since
8-bit finishing recovers as much as full precision (\S\ref{sec:return}). The cap and the number of finishing steps are
chosen per budget on development rows and scored on held-out rows (Appendix~\ref{app:controller-finishing}). The
controller costs less than stopping alone because the development rows pick a lower cap once finishing is available;
finishing recovers what extra compressed loops would.

\paragraph{Cost is weight traffic.}
At batch size one a looped model reads its whole weight set on every loop, and that read sets the time of the loop. We
charge each loop by the bytes of weights it reads. A compressed loop costs one unit, and an 8-bit finishing
step costs two to four (Appendix~\ref{app:protocol-cost}).

\paragraph{Sampling is an optional add-on.}
A third lever runs several perturbed trajectories and keeps the one the head trusts most. Here it perturbs the latent
state, as in PTRM \citep{sghaier2026ptrm}; perturbing the rounding itself is studied in
Appendix~\ref{app:controller-sampling}.

\begin{algorithm}[t]
\caption{The controller: stop when the halting head fires, then finish with an 8-bit copy of the weights.}
\label{alg:controller}
\small
\begin{algorithmic}[1]
\Require compressed weights $\tilde\theta$, an 8-bit copy $\theta_8$, depth cap $D$, finishing steps $j$, input $x$
\State $z \gets z_0$
\For{$t = 1, \dots, D$}
  \State $z \gets f_{\tilde\theta}(z, x)$ \Comment{one compressed loop; cost 1}
  \If{the halting head fires on $z$} \textbf{break} \EndIf
\EndFor
\For{$j$ steps} $z \gets f_{\theta_8}(z, x)$ \Comment{finishing steps; cost $8/b_{\text{eff}}$ each}
\EndFor
\State \Return $g_{\theta_8}(z)$ \Comment{readout from the 8-bit copy; no full-precision weights}
\end{algorithmic}
\end{algorithm}

\subsection{Results}
\label{sec:controller-results}

\begin{table}[t]
\caption{\textbf{Stopping at the halting head and finishing with 8-bit loops beats fixed depth at under a third of
the weight traffic.} Held-out exact accuracy on Sudoku-Extreme, averaged over 20 compressed models; cost in
compressed loops per puzzle; finishing steps per puzzle. The first block adds one lever at a time. The second block
holds the alternatives: uniform refinement finishes every puzzle and confidence-gated refinement finishes the
puzzles whose halting logit is low, both at full precision. The last block tests the head's choice of puzzles to
finish against a random set of the same size. Right, the same rows on the cost--accuracy plane; the path is the
first block, the dotted step the add-on, the open grey points the alternatives. Paired 95\% intervals; protocol in
App.~\ref{app:controller}.}
\label{tab:controller}
\begin{minipage}[c]{0.58\textwidth}
\centering\scriptsize
\setlength{\tabcolsep}{2.5pt}
\begin{tabular}{@{}lrrr@{}}
\toprule
Method & Accuracy (\%) & Cost & Finishing steps \\
\midrule
Fixed depth                                        & 61.1 & 55.0 & 0    \\ 
\quad + stop at the halting head                   & 63.0 & 23.4 & 0    \\ 
\rowcolor{sigrow}\qquad + 8-bit finishing = \textbf{controller}     & \textbf{75.8} & \textbf{16.1} & 1.13 \\ 
\qquad controller $-$ fixed depth                  & \multicolumn{3}{@{}l}{$+14.7$ [$+14.3$, $+15.2$]} \\ 
\qquad\quad + latent sampling (add-on)             & 78.7 & 28.4 & 1.45 \\ 
\midrule
Controller with full-precision finishing           & 75.7 & 25.8 & 1.20 \\ 
Uniform refinement, full precision                 & 72.4 & 48.3 & 4.00 \\ 
Confidence-gated refinement, full precision        & 71.5 & 27.6 & 2.09 \\ 
\midrule
Finish where the head has not stopped              & 73.2 & 10.6 & --   \\ 
Finish a random set of the same size               & \multicolumn{3}{@{}l}{71.3 [71.2, 71.4]\ \ same cost} \\ 
\bottomrule
\end{tabular}
\end{minipage}\hfill
\begin{minipage}[c]{0.41\textwidth}
\centering
\includegraphics[width=\linewidth]{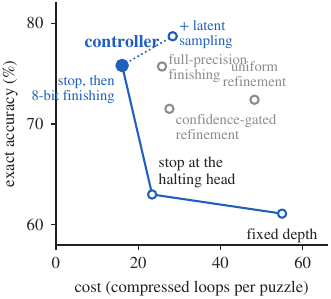} 
\end{minipage}
\end{table}

\paragraph{Stopping and finishing lift accuracy by 15 points at under a third of the traffic.}
On the 20 compressed Sudoku-Extreme models, fixed-depth inference scores 61.1 at a cost of 55 compressed loops per
puzzle. Stopping at the halting head alone scores 63.0 at a cost of 23.4, so the head already rescues answers that
fixed depth loops past. Finishing with 8-bit weights then lifts accuracy to 75.8 at a cost of 16.1, a gain of 14.7
points [14.3, 15.2] over fixed depth (Table~\ref{tab:controller}, and its right panel). On the four Maze-Hard models
the same controller lifts fixed depth from 70.9 to 78.5 at the same cost (Appendix~\ref{app:controller-cells}).
The gain sits on the
collapsed models, as their return predicts (\S\ref{sec:return}). Models that survive at the default depth keep their
accuracy at a fraction of the cost, so for a model that never failed the controller is an efficiency gain. Per model,
the controller is above fixed depth with an interval above zero in 10 of the 20 and below it in 4, by 0.4 to 2.4
points, all of them surviving models where stopping alone was chosen (Appendix~\ref{app:controller-cells}). The
sampling add-on buys three more points at nearly twice the cost, and with it
every model ends above fixed depth.

\paragraph{The alternatives fall short on both axes.}
Uniform refinement, which finishes every puzzle at full precision, reaches 72.4 at three times the controller's cost,
and gating that refinement on the head's confidence reaches 71.5 at 27.6 (Table~\ref{tab:controller}, second block).
The gain also survives a tighter budget. At 24 cost units the controller with full-precision finishing scores 72.2
against 60.5 for fixed depth (Appendix~\ref{app:controller-cells}).

\paragraph{No full-precision copy is needed.}
Finishing with 8-bit weights reaches the accuracy of finishing with full-precision weights, 75.8 against 75.7, at a
cost of 16.1 against 25.8, since an 8-bit step reads a quarter of the bytes (\S\ref{sec:controller-levers}). A
deployed model stores its
compressed weights and an 8-bit copy of them.

\paragraph{The halting head knows which puzzles to finish.}
Finishing could help regardless of which puzzles are finished. To check, we hold the number of finished puzzles and
the cost fixed and finish a random set instead of the puzzles the head has not stopped. The head's choice scores
73.2 against 71.3 for random sets, and it is ahead in every model where the two differ
(Appendix~\ref{app:controller-random}). The head's firing carries information about which puzzles still need
finishing.

\paragraph{The halting head is the better stopping rule.}
A threshold on the size of the latent update, frozen on full precision, stops earlier than the head and matches fixed
depth, but it falls below the head in 22 of 46 models, among them several whose trajectories pass the answer before
they settle (Appendix~\ref{app:controller-tab2}). Where no halting head exists, as in implicit models and looped
language models, it is the available rule.

For a deployer, the recipe is to stop when the model says it is done and to finish the rest with an 8-bit copy; the
compressed model never needs full-precision weights at test time.
\section{Related work}
\label{sec:related}

\paragraph{Looped and recurrent-depth models.}
Weight-tied depth runs from Universal Transformers \citep{dehghani2019universal} through looped transformers
\citep{giannou2023looped,yang2024looped,saunshi2025latent}, looped language models
\citep{bae2025relaxed,bae2025mor,geiping2025recurrentdepth,zhu2025ouro,mcleish2025retrofitted,hao2024coconut},
equilibrium models \citep{bai2019deq,bai2020mdeq} and recurrent puzzle solvers
\citep{schwarzschild2021deepthinking,bansal2022overthinking,wang2025hrm,jolicoeur2025trm}. A loop that settles turns
extra iterations into accuracy \citep{anil2022pathindependence}; we find that settling also decides what compression
costs.

\paragraph{Compressing looped models.}
LoopQ treats recursive error accumulation as the obstacle and proposes a loop-aware quantizer \citep{fang2026loopq};
edge-deployment work on TRM states the same account \citep{jim2026survives}; concurrent work on TRM-style reasoners
attributes their collapse to an accumulating bias \citep{ingolfsson2026quantizing}. We show that collapse needs no
bias, that the collapsed loop still settles, and that the lost answers can be recovered. For monotone equilibrium
networks, quantization provably preserves convergence and bounds the shift
\citep{mondeqquant2026,winston2020mondeq}; our reasoners carry no such certificate, and we measure the same behaviour
in them. Quantized reasoning language models, whose loop runs through generated tokens, gain less from
longer reasoning \citep{liu2025quanthurts,lotfi2026quantizedthink}; our loops run in the latent state.

\paragraph{Perturbation, refinement and halting.}
The picture transfers classical results to learned loops: the fixed-point perturbation bound
\citep{banach1922operations}, the margin-over-sensitivity radius \citep{tsuzuku2018lipschitz}, mixed-precision
iterative refinement \citep{carson2018threeprecisions,higham2022mixed}, whose accuracy analogue is our finishing law,
and stochastic rounding, whose unbiased errors average out \citep{gupta2015limited,connolly2021stochastic} as our
fresh error does inside a loop. Jacobian regularization of equilibrium models \citep{bai2021jacobian} is a
training-time route to lower sensitivity. Adaptive computation \citep{graves2016act,banino2021pondernet} and
confidence-based early exit \citep{schuster2022calm} learn when to stop, and overthinking names the answers that later
loops lose \citep{kaya2019shallowdeep,bansal2022overthinking}; the halting heads of TRM and HRM already separate right
from wrong answers in full precision \citep{sghaier2026ptrm,haltingverifier2026a}. Our controller uses the same head
under compression and adds finishing steps, which early exit alone lacks.
\section{Discussion and conclusion}
\label{sec:conclusion}

\paragraph{Compression and depth stop being rivals.}
Once a loop is known to settle, the error that compression adds is paid once, and a deployer can compress the block and
let the model think as long as the budget allows. The push against the room, measured without labels, says which
formats survive (\S\ref{sec:room}), and a few 8-bit finishing steps buy back what a coarse format loses, at the end of
the computation, where the halting head says the model is done (\S\ref{sec:controller}). A loop that does not settle
is outside the picture; in deterministic samplers the error grows with depth as the usual account says.

\paragraph{Training decides the room and the push.}
Both move. Attention mixers have several times the room of MLP mixers, and quantization-aware training lowers the
sensitivity while leaving the room where it was (\S\ref{sec:room}), so a designer who wants a compressible looped model
trains the loop to settle, chooses the mixer with the wider room, and keeps a halting head, which stays a reliable
signal under compression. Penalties on the loop's sensitivity from the equilibrium-model literature
\citep{bai2021jacobian} are the natural next step.

\paragraph{Limitations.}
The largest looped model we compress has a few billion parameters, and we cannot say whether much larger ones settle.
Our costs are modelled from weight traffic and not timed on hardware.
The depth claim covers supervision steps and outer cycles, since some series widen along the fast inner cycles, and our
comparison of fixed and fresh error at depth places the fixed error in the weights and the fresh error in the
activations. Under a fixed error the attention mixer sometimes repairs its gap, for reasons we have not found, and the
one room shared across models over-predicts MLP mixers and under-predicts attention. Ouro does not settle, so the
picture says little about it.

\paragraph{Conclusion.}
Compression and depth have been treated as rivals in looped models, and we set out expecting to measure how fast the
rounding error piles up. It does not pile up. For a loop that settles, a fixed rounding error tilts the bowl once,
thinking longer moves the answer no further, the tilt can be measured before deployment, and a few precise loops undo
it afterwards, with tools that numerical analysis has used for decades. We built the controller to test that reading
at deployment scale, and it turns collapsed low-bit reasoners into models that beat their own fixed-depth inference at
a third of the weight traffic. A looped model reasons in a latent state that no one reads, which makes its reasoning
harder to monitor than a written chain of thought \citep{korbak2025monitorability}; fidelity to the full-precision
state is the cheap, label-free check that a compressed model still settles where the original does. The test for a
new looped model costs one run. Loop it longer and watch whether the gap grows.

\section*{AI use disclosure}
We used AI coding and writing assistants in this work. Under the authors' direction, they helped implement code, run
and monitor experiments, and edit the prose of this paper. The research ideas, the hypotheses,
the choice of experiments and the direction of the work are the authors'. Every number in the paper was checked against the written result
records, and the authors take full responsibility for the paper's content.

\section*{Reproducibility statement}
We will release the code, the compressed and trained checkpoints, and every result file that the tables and figures are
built from. Models, tasks and training are described in App.~\ref{app:models}; the compression formats, the evaluation
protocol, the statistics, the platforms and the cost model in App.~\ref{app:protocol}. Every comparison pairs a compressed model with its own full-precision reference on the same
examples and the same platform.
The code, the checkpoints and the result files will be released with the camera-ready version.

\ifx\buildmode\buildsubmission\else
\fi

\bibliography{custom,sections/paper/bib_a1,sections/paper/bib_a2,sections/paper/bib_a3,sections/paper/bib_a4}
\bibliographystyle{iclr2027_conference}

\appendix
\clearpage
\etocdepthtag.toc{appendixtag}
\etocsettagdepth{maintag}{none}
\etocsettagdepth{appendixtag}{subsection}
\begingroup
\hypersetup{pdfborder={0 0 0}}
\etocsettocstyle{\noindent{\large\bfseries Appendix contents}\par\vspace{0.8em}}{\vspace{1.2em}}
\tableofcontents
\endgroup

\section{Models}
\label{app:models}

The paper compresses 33 models from five families (Table~\ref{tab:app-models}): five released puzzle reasoners,
19 TRMs we train at several widths and seeds, one DeepThinking maze solver evaluated on two maze sizes, three implicit
models, four looped language models and one diffusion model, plus quantization-aware-trained variants
(Appendix~\ref{app:widthlaw}). Every exhibit compares a compressed model with its own full-precision version on the same
examples, so a checkpoint's absolute accuracy enters no claim. That matters for the released TRM Sudoku-MLP
checkpoint, a community reproduction that scores below the original paper.

\begin{table}[!htbp]
\caption{\textbf{Every model, the map it repeats, its default depth and its metric; 33 models in all.} ``What loops'' is the
weight-tied map; ``default depth'' is the released depth; $n$ is the number of evaluation examples per compressed
model in the main experiments of \S\ref{sec:shift}.}
\label{tab:app-models}
\centering\scriptsize
\begin{tabular}{@{}p{0.17\textwidth}r p{0.22\textwidth}p{0.17\textwidth}p{0.14\textwidth}p{0.12\textwidth}@{}}
\toprule
Model & \# & What loops & Default depth & Task, metric & $n$ \\
\midrule
TRM, MLP mixer (released) & 1 & one 2-layer block, reused for the $z_L$ and $z_H$ updates; token-mixing SwiGLU & $H{=}3$, $L{=}6$, 16 supervision steps & Sudoku-Extreme (97 positions), exact & 1000; 1800 (depth to 512) \\
TRM, attention mixer (released) & 2 & as above, self-attention with RoPE; one checkpoint per task & as above & Sudoku-Extreme; Maze-Hard $30{\times}30$ (916 positions), exact and valid path & 1000 / 500 \\
HRM (released) & 2 & separate high- and low-level modules (4 layers each); one checkpoint per task & $H{=}2$, $L{=}2$, 16 supervision steps & Sudoku-Extreme, Maze-Hard & 1000 / 500 \\
Our TRMs, MLP mixer & 10 & as TRM; width 256 (seeds 0--2), 512 (seeds 0--5), 768 (seed 0) & as TRM & Sudoku-Extreme, exact & 1000 \\
Our TRMs, attention mixer & 9 & as TRM; width 256 (seeds 0--2), 512 (seeds 0--5) & as TRM & Sudoku-Extreme, exact & 1000 \\
DeepThinking & 1 & recall block, input concatenated at every iteration (width 128) & 300 iterations; swept to 1000 & mazes of size 13 and 33, exact & per size \\
MDEQ-Small, -XL & 2 & multiscale equilibrium block, Broyden solver & 26 solver iterations & ImageNet, top-1 & 5000 / 2000 \\
DEQ-Transformer & 1 & equilibrium transformer layer, Anderson solver & swept to 60 & WikiText-103, perplexity & full test set \\
Huginn-3.5B, Recurrent-OLMo-2 & 2 & core block between a fixed prelude and coda & set per call & GSM8K, greedy accuracy & Appendix~\ref{app:lms} \\
Ouro-1.4B, -2.6B & 2 & the whole layer stack, with a trained exit gate & 4 loops & GSM8K & Appendix~\ref{app:lms} \\
DDPM UNet & 1 & the denoiser, one call per sampler step & DDPM 50--1000, DDIM 10--100 steps & CIFAR-10, FID-5k to full-precision samples & 5000 samples \\
\midrule
Total & 33 & & & & \\
\bottomrule
\end{tabular}
\end{table}
\paragraph{Loops in each family.} TRM keeps two latent states and applies one two-layer block $L$ times to update
$z_L$, then once to update $z_H$; $H$ such cycles make one supervision step \citep{jolicoeur2025trm}. HRM has the same
structure with two separate modules \citep{wang2025hrm}. A linear head reads the answer from $z_H$, and the halting head
reads the first position of $z_H$. DeepThinking applies a convolutional recall block to the current state and the input
at every iteration \citep{bansal2022overthinking}. The DEQs run Broyden (MDEQ) or Anderson (DEQ-Transformer) solvers,
which we re-implement with per-example tracking of the lowest-residual iterate \citep{bai2019deq,bai2020mdeq}. The
looped language models loop a core block between a fixed prelude and coda (Huginn, Recurrent-OLMo-2) or loop the whole
stack with a trained exit gate (Ouro). For diffusion, every compressed model starts from the same noise as its
full-precision twin and, under DDPM, draws the same step noise.

\paragraph{Training our TRMs.} We train with the upstream recipe: global batch 768, learning rate $10^{-4}$ with 2000
warm-up steps, weight decay 1.0, EMA 0.999, $H{=}3$, $L{=}6$, two layers, 16 supervision steps and 65{,}104 steps on
the 1000-puzzle augmented Sudoku-Extreme training set. The upstream trainer's resume bug is fixed in ours, and no run
resumed. 

\FloatBarrier

\section{Compression, statistics and cost}
\label{app:protocol}

\subsection{Formats and injected error}
\label{app:protocol-quant}

Symmetric round-to-nearest computes $\tilde W = s\cdot\mathrm{clip}(\mathrm{round}(W/s), -q, q)$, $q = 2^{b-1}-1$,
with $s = \max|W|/q$ over one tensor (t), one output row (c), or 32 or 128 consecutive input weights (g32, g128); the
asymmetric per-channel format (a) uses the row's min--max range and a zero point. Quantization is simulated in floating
point. Every linear layer of TRM and HRM is quantized, heads included, and embeddings stay at full precision; the other
families quantize every convolution and linear layer of the looped block. The suffix +a8 adds static per-tensor INT8
activations. For the comparison with concurrent work we add its W4A4 formats, with per-tensor activation scales clipped
either at the maximum over all calibration calls or at the mean per-call 99.99th percentile, and MXInt4
\citep{rouhani2023microscaling}. Calibration rows never overlap evaluation rows. 

A \emph{fixed error} adds $\sigma\,\mathrm{std}(W)\,\epsilon$ to each weight tensor once, with $\epsilon\sim\mathcal
N(0,I)$. A \emph{fresh error} adds $\sigma\,\mathrm{std}(a)\,\epsilon$ to the input $a$ of every linear layer at every
call, with a new draw. Appendix~\ref{app:fixedfresh-arms} also uses three variants of the concurrent work
\citep{ingolfsson2026quantizing}: activation noise drawn once and reused at every call, a sign-consistent activation
bias (80\% of its norm in its mean), and a sign-consistent weight bias, all at matched per-call Frobenius norm. The
per-call output error $\rho = \lVert y - W_0 x\rVert/\lVert W_0 x\rVert$ is measured along each model's own trajectory.

\subsection{Pairing and verdicts}
\label{app:protocol-stats}

Every compressed model runs with its own full-precision reference on the same examples, at the same depth, on the same
platform. The two platforms are nodes with L40S cards, platform (i) in the figures, and nodes with A100 (40\,GB) cards,
platform (ii); their software stacks differ, and no number from one is compared with or pooled with a number from
the other, except where Appendix~\ref{app:lms_detect} says so. A \emph{series} is a model, a format and
one loop dial. We compute a 95\% paired bootstrap interval over examples (2000 resamples) for the change of the gap from
the shallowest to the deepest point: the series \emph{widens} if the interval lies below zero, \emph{narrows} if above,
and is \emph{flat} otherwise \citep{efron1993bootstrap}. A second verdict, on the slope of the gap against $\log_2$ of
the loops, catches widening between the endpoints. For the runs to 512 supervision steps (Appendix~\ref{app:fixedfresh}),
with $D = \mathrm{gap}(512) - \mathrm{gap}(16)$: \emph{catch-up} if $D>0$, full if the interval of gap(512) includes
zero; \emph{widening} if $D<0$; \emph{tracks} if $D$ and gap(512) both include zero; \emph{constant gap} if $D$
includes zero and gap(512) does not. 

\subsection{Pruning and distillation erase the puzzle reasoners}
\label{app:other-compression}

Quantization is the compression these models survive. Structured magnitude pruning at 25\% sparsity already leaves
every puzzle reasoner unable to solve a single puzzle, and so does distillation into a one-layer, one-cycle student of
width 256 trained with a KL objective (Table~\ref{tab:app-prune-distill}). Both keep much of the per-cell accuracy on
Maze and Sudoku and lose every whole puzzle.

\begin{table}[h]
\centering
\caption{\textbf{Pruning and distillation leave the puzzle reasoners unable to solve a single puzzle.} Exact puzzle
accuracy and cell accuracy (\%) of the dense model, after structured pruning, and of a distilled student (Maze
teacher 6.82M, student 855K parameters; Sudoku teacher 5.03M, student 745K). Released checkpoints at $H{=}3$ with 10
supervision steps (ARC: 16 steps with test-time augmentation), full test sets, one run.} 
\label{tab:app-prune-distill}
\small
\begin{tabular}{@{}lrrrrrrrr@{}}
\toprule
 & \multicolumn{2}{c}{Dense} & \multicolumn{2}{c}{25\% pruned} & \multicolumn{2}{c}{50\% pruned} & \multicolumn{2}{c}{Distilled} \\
\cmidrule(lr){2-3}\cmidrule(lr){4-5}\cmidrule(lr){6-7}\cmidrule(lr){8-9}
Task & exact & cell & exact & cell & exact & cell & exact & cell \\
\midrule
ARC    & 36.0 & 87.6 & 0.0 & 0.3  & 0.0 & 0.6  & --  & --   \\ 
Maze   & 86.8 & 99.5 & 0.0 & 86.4 & 0.0 & 0.0  & 0.0 & 87.3 \\ 
Sudoku & 69.1 & 87.5 & 0.0 & 50.0 & 0.0 & 37.8 & 0.0 & 54.8 \\ 
\bottomrule
\end{tabular}
\end{table}
\subsection{Weight-traffic cost model}
\label{app:protocol-cost}

At batch size one, a loop of a small looped model is bound by reading its weights, so the controller's cost counts the
bits of weights read. One compressed loop reads $b_{\text{eff}}$ bits per weight, scales included; a full-precision loop
reads 32 and costs $r = 32/b_{\text{eff}}$ compressed loops (Table~\ref{tab:app-cost}), and an 8-bit finishing step costs
$8/b_{\text{eff}}$. Activations and compute are not counted, and no cost is a measured time. 

\begin{table}[!htbp]
\caption{\textbf{One full-precision loop costs 4 to 16 compressed loops.} $r = 32/b_{\text{eff}}$, with
$b_{\text{eff}}$ the bits read per weight including group scales. Weight traffic only.}
\label{tab:app-cost}
\centering\small
\begin{tabular}{@{}lcccccc@{}}
\toprule
Compressed format & w4 & w3 & w3g32 & w2 & w2g32 & w8 \\
\midrule
$b_{\text{eff}}$ & 4 & 3 & 3.5 & 2 & 2.5 & 8 \\
$r$ (full-precision loop) & 8 & 10.7 & 9.1 & 16 & 12.8 & 4 \\
\bottomrule
\end{tabular}
\end{table}
\FloatBarrier

\section{Thinking longer}
\label{app:depth}

Figure~\ref{fig:teaser}b normalises each family's loop axis by its default depth (Table~\ref{tab:app-models}) and
plots the change in the gap from the default depth on, with a paired bootstrap interval of that change over evaluation
examples (2000 resamples), so that a model that is still settling at shallow depth is not read as an effect of looping
longer; Figure~\ref{fig:teaser}c plots FID on the same axis. Each series comes from one platform.

\subsection{No surviving puzzle reasoner loses accuracy along outer cycles}
\label{app:depth-series}

Across four puzzle reasoners and six formats, no surviving series widens along outer cycles in any of three runs, and
along supervision steps only per-tensor 4-bit weights ever widen (Tables~\ref{tab:app-depth-summary}
and~\ref{tab:app-depth-p4d}). The runs are the A100 platform, the L40S platform with the same rows (a hardware and
library replicate), and the L40S platform with new rows and a new calibration draw (an independent replicate). Along the
fast inner cycles three A100 series widen, so we make no claim for that axis, which is the one the concurrent work
varies \citep{ingolfsson2026quantizing}. Series that collapse do widen, but only because full precision improves while
the collapsed model stays near zero: Sudoku-MLP w4t goes from 2.1\% to 5.7\% between 1 and 32 supervision steps while
full precision goes from 28.4\% to 74.2\%.

\begin{table}[!htbp]
\caption{\textbf{Along supervision steps and outer cycles the gap of a surviving model almost never widens.}
Number of surviving series that widen / stay flat / narrow, by endpoints and by trend (95\% paired bootstrap,
2000 resamples). Sudoku $n{=}1000$, Maze $n{=}500$. Each row comes from one platform. The L40S replicate did not
sweep $L$; its Maze $H$ series cover $H$ 1--3 in the audited records, and records to deeper $H$ leave every verdict
unchanged.}
\label{tab:app-depth-summary}
\centering\small
\begin{tabular}{@{}l ccc ccc@{}}
\toprule
& \multicolumn{3}{c}{Endpoints} & \multicolumn{3}{c}{Trend} \\
\cmidrule(lr){2-4}\cmidrule(l){5-7}
Run (series) & $n_{\text{sup}}$ & $H$ & $L$ & $n_{\text{sup}}$ & $H$ & $L$ \\
\midrule
A100 (22) & 1 / 16 / 5 & 0 / 18 / 4 & 3 / 14 / 5 & 0 / 15 / 7 & 0 / 19 / 3 & 5 / 12 / 5 \\
L40S replicate (23) & 0 / 15 / 8 & 0 / 20 / 3 & -- & 0 / 18 / 5 & 0 / 20 / 3 & -- \\
L40S independent (23) & 2 / 18 / 3 & 0 / 21 / 2 & -- & 1 / 18 / 4 & 0 / 21 / 2 & -- \\
\bottomrule
\end{tabular}
\end{table}
\begin{table}[!htbp]
\caption{\textbf{A100 platform: no surviving series widens along outer cycles, and one widens along supervision
steps.} Change of the gap from the shallowest to the deepest point, in points of exact accuracy [95\% paired
bootstrap interval]; $\uparrow$ widens, $=$ flat, $\downarrow$ narrows (endpoints). $n_{\text{sup}}$ 1$\to$32; $H$
1$\to$6 (HRM 1$\to$4); $L$ 1$\to$8 (HRM 1$\to$4); default $H$, $L$ and 16 supervision steps on the other axes.
Sudoku $n{=}1000$, Maze $n{=}500$.}
\label{tab:app-depth-p4d}
\centering\scriptsize
\begin{tabular}{@{}ll lll@{}}
\toprule
Model & Format & $n_{\text{sup}}$ & $H$ & $L$ \\
\midrule
TRM Sudoku, MLP & w4c & +1.1 [$-$1.9, +4.1] = & $-$1.2 [$-$3.3, +1.1] = & +1.4 [$-$1.9, +4.8] = \\
 & w4g32 & $-$0.2 [$-$2.4, +2.1] = & $-$1.2 [$-$3.3, +0.9] = & $-$1.3 [$-$4.4, +1.8] = \\
 & W4A4 g32 & +5.0 [+2.2, +7.9] $\downarrow$ & $-$0.3 [$-$2.5, +1.9] = & $-$7.5 [$-$10.7, $-$4.5] $\uparrow$ \\
 & W4A4 MXInt4 & +7.5 [+4.1, +10.9] $\downarrow$ & +0.1 [$-$2.3, +2.3] = & $-$11.3 [$-$14.4, $-$7.9] $\uparrow$ \\
 & W8A8 per-tensor & +13.8 [+10.8, +16.9] $\downarrow$ & +2.5 [+0.2, +5.0] $\downarrow$ & +7.3 [+3.8, +10.8] $\downarrow$ \\
TRM Sudoku, attn. & w4c & +0.7 [$-$1.0, +2.4] = & +0.3 [$-$2.0, +2.6] = & +4.7 [+2.3, +7.1] $\downarrow$ \\
 & w4g32 & +1.0 [$-$0.7, +2.6] = & +0.4 [$-$2.0, +2.7] = & +3.1 [+0.8, +5.4] $\downarrow$ \\
 & W4A4 g32 & +1.9 [+0.0, +3.6] = & +0.5 [$-$2.0, +3.0] = & +5.3 [+2.9, +7.8] $\downarrow$ \\
 & W4A4 MXInt4 & +1.4 [$-$0.6, +3.3] = & +3.5 [+1.0, +6.1] $\downarrow$ & +0.7 [$-$1.9, +3.5] = \\
 & w4t & $-$2.8 [$-$5.0, $-$0.5] $\uparrow$ & +1.7 [$-$1.1, +4.3] = & $-$5.5 [$-$8.8, $-$2.3] $\uparrow$ \\
 & W8A8 per-tensor & +0.7 [$-$1.2, +2.5] = & +0.4 [$-$2.0, +2.8] = & $-$2.5 [$-$5.1, +0.3] = \\
TRM Maze, attn. & w4c & $-$0.8 [$-$2.0, +0.4] = & +0.0 [+0.0, +0.0] = & +0.0 [$-$1.4, +1.4] = \\
 & w4g32 & +0.0 [$-$1.2, +1.2] = & $-$0.2 [$-$0.6, +0.0] = & $-$1.0 [$-$2.4, +0.2] = \\
 & W4A4 g32 & +0.0 [$-$1.8, +2.0] = & +0.0 [+0.0, +0.0] = & +1.0 [$-$0.6, +2.8] = \\
 & W4A4 MXInt4 & +2.2 [$-$0.2, +4.6] = & $-$0.2 [$-$0.6, +0.0] = & +1.4 [$-$0.8, +3.4] = \\
 & w4t & +1.4 [$-$1.0, +4.0] = & +0.0 [$-$0.6, +0.6] = & +0.2 [$-$2.0, +2.4] = \\
HRM Sudoku & w4c & +0.6 [$-$1.1, +2.2] = & +0.7 [$-$1.1, +2.5] = & +1.0 [$-$1.7, +3.6] = \\
 & w4g32 & +0.7 [$-$0.9, +2.3] = & +0.3 [$-$1.4, +1.8] = & +1.2 [$-$1.0, +3.4] = \\
 & W4A4 g32 & +2.5 [+0.7, +4.3] $\downarrow$ & +2.5 [+0.6, +4.4] $\downarrow$ & +1.6 [$-$1.0, +4.2] = \\
 & W4A4 MXInt4 & +0.9 [$-$1.1, +2.7] = & +1.5 [$-$0.5, +3.5] = & +2.6 [$-$0.1, +5.5] = \\
 & w4t & $-$0.6 [$-$2.5, +1.3] = & +1.7 [$-$0.3, +3.7] = & +2.8 [+0.1, +5.5] $\downarrow$ \\
 & W8A8 per-tensor & +2.5 [+0.7, +4.2] $\downarrow$ & +2.0 [+0.1, +3.8] $\downarrow$ & +1.2 [$-$1.2, +3.7] = \\
\bottomrule
\end{tabular}
\end{table}
\FloatBarrier

\subsection{Implicit models and recurrent convnets lose nothing to extra loops while they survive}
\label{app:depth-deq}

Once its solver converges, MDEQ-Small changes by at most 0.1 points from 26 to 100 Broyden iterations, for every noise
level and format (Table~\ref{tab:app-mdeq}). The DEQ-Transformer improves with extra Anderson iterations for as long as
its perplexity stays within twice full precision's, and worsens only after it has collapsed (Table~\ref{tab:app-deqlm}).
Its collapsed w4t model converges to a residual 481 times smaller than full precision's, so a residual check would pass
it. DeepThinking maze solvers change by at most 1.1 points when their iterations double from 500 to 1000.

\begin{table}[!htbp]
\caption{\textbf{MDEQ-Small on ImageNet: accuracy stops changing once the solver converges, at every noise
level, and the solver converges on collapsed models too.} Top-1 \% by Broyden budget $k$ (26 = default);
cosine to the full-precision equilibrium; fraction of images with residual below $10^{-2}$ at $k{=}100$; median
steps to $10^{-2}$. Noise rows: mean [min, max] over 3 noise directions at $k{=}26$; quantizer rows give the mean
relative weight error. $n{=}5000$ validation images; L40S platform.}
\label{tab:app-mdeq}
\centering\scriptsize
\begin{tabular}{@{}lllllll@{}}
\toprule
Noise / format & $k{=}8$ & $k{=}16$ & $k{=}26$ & $k{=}100$ & Cosine to full precision & Converged / median steps \\
\midrule
fp32 & 72.2 & 73.9 & 73.8 & 73.8 & 1 & 99.7\% / 24 \\
$\sigma$ 0.05 & 71.4 & 72.9 & 73.1 [72.7, 73.7] & 73.0 & 0.978 & 99.7\% / 24 \\
$\sigma$ 0.1 & 67.3 & 69.0 & 68.9 [67.1, 70.7] & 68.9 & 0.923 & 99.8\% / 24 \\
$\sigma$ 0.2 & 46.0 & 49.1 & 49.1 [47.1, 52.9] & 49.1 & 0.779 & 99.8\% / 25 \\
$\sigma$ 0.3 & 12.6 & 14.6 & 14.8 [10.5, 18.4] & 14.8 & 0.614 & 99.8\% / 25 \\
$\sigma$ 0.5 & 0.3 & 0.2 & 0.3 & 0.3 & 0.334 & 99.5\% / 26 \\
w8c (0.009) & 72.2 & 74.0 & 73.8 & 73.8 & 0.999 & 99.7\% / 24 \\
w4g32 (0.107) & 64.8 & 66.6 & 66.6 & 66.5 & 0.882 & 99.6\% / 26 \\
w4c (0.171) & 56.2 & 58.0 & 57.8 & 57.9 & 0.850 & 99.5\% / 25 \\
w4t (0.408) & 0.2 & 0.2 & 0.2 & 0.2 & 0.540 & 99.9\% / 19 \\
\bottomrule
\end{tabular}
\end{table}
\begin{table}[!htbp]
\caption{\textbf{DEQ-Transformer on WikiText-103: extra iterations help as long as the model survives and hurt only
after it has collapsed; the collapsed quantizer has the smallest residual.} Test perplexity by Anderson
iteration cap $k$ (lower is better), median relative residual and cosine to full precision at $k{=}60$. Noise rows:
mean [min, max] over 3 noise directions at $k{=}60$; quantizer rows give the relative weight error. Full test set
(245{,}569 tokens); L40S platform.}
\label{tab:app-deqlm}
\centering\scriptsize
\begin{tabular}{@{}llllll@{}}
\toprule
Noise / format & $k{=}10$ & $k{=}30$ & $k{=}60$ & Residual, $k{=}60$ & Cosine, $k{=}60$ \\
\midrule
fp32 & 91.2 & 24.48 & 23.14 & 2.5e-2 & 1 \\
$\sigma$ 0.1 & 92.3 & 25.23 & 23.84 [23.82, 23.86] & 2.4e-2 & 0.981 \\
$\sigma$ 0.3 & 98.5 & 34.59 & 32.65 [32.63, 32.70] & 2.3e-2 & 0.846 \\
$\sigma$ 0.5 & 219 & 124.7 & 127.1 [124.2, 132.0] & 2.9e-2 & 0.55 \\
$\sigma$ 0.7 & 923 & 1071 & 1182 [1142, 1204] & 6.1e-2 & 0.29 \\
w8c (0.008) & 90.2 & 24.48 & 23.14 & 2.5e-2 & 1.000 \\
w4g32 (0.09) & 96.8 & 25.11 & 23.74 & 2.5e-2 & 0.984 \\
w4c (0.15) & 67.1 & 25.61 & 24.35 & 2.0e-2 & 0.956 \\
w4t (0.61) & 2556 & 2401 & 2395 & 5.2e-5 & 0.27 \\
\bottomrule
\end{tabular}
\end{table}
\FloatBarrier

\subsection{The deterministic sampler compounds error with steps; the ancestral sampler does not}
\label{app:depth-diffusion}

DDPM and DDIM share one denoiser; DDPM draws fresh noise at every step and DDIM is deterministic. At 4-bit grouped
weights, DDPM's distance to the full-precision samples is the same from 50 to 1000 steps, whereas DDIM's grows from 39.5
at 10 steps to 304 at 100 (Table~\ref{tab:app-ddpm}). Outside DDPM's tolerance, its cost also grows with steps.

\begin{table}[!htbp]
\caption{\textbf{The deterministic DDIM sampler compounds with steps; ancestral DDPM keeps a step-flat cost inside
its tolerance.} FID-5k against the full-precision samples of the same sampler, drawn from the same starting and
step noise (lower is closer). Noise rows: mean over 3 noise directions; quantizer rows give the relative weight
error. DDPM CIFAR-10 UNet; L40S platform. The quantizer rows of the DDPM-1000 column need no noise draw.}
\label{tab:app-ddpm}
\centering\scriptsize
\begin{tabular}{@{}l rrrr rrrr@{}}
\toprule
& \multicolumn{4}{c}{DDIM steps} & \multicolumn{4}{c}{DDPM steps} \\
\cmidrule(lr){2-5}\cmidrule(l){6-9}
$\sigma$ / format & 10 & 25 & 50 & 100 & 50 & 100 & 250 & 1000 \\
\midrule
$\sigma$ 0.01 & 7.4 & 6.6 & 6.1 & 6.7 & 1.5 & 2.1 & 2.6 & -- \\
$\sigma$ 0.03 & 26.5 & 17.0 & 17.2 & 49.8 & 6.2 & 7.1 & 7.9 & -- \\
$\sigma$ 0.05 & 45.7 & 21.2 & 61.4 & 189.5 & 12.7 & 13.3 & 13.8 & -- \\
$\sigma$ 0.1 & 95.6 & 136.5 & 345.8 & 397.6 & 43.2 & 42.8 & 42.1 & 46.1 \\
$\sigma$ 0.3 & 386 & 358 & 356 & 312 & 213 & 255 & 290 & -- \\
w8c (0.010) & 6.4 & 5.4 & 5.0 & 6.0 & 2.0 & 2.6 & 3.2 & -- \\
w4g32 (0.105) & 39.5 & 49.5 & 164 & 304 & 24.3 & 23.5 & 22.1 & 22.3 \\
w4c (0.179) & 205 & 367 & 396 & 375 & 116 & 139 & 159 & 189 \\
w4t (0.369) & 405 & 407 & 382 & 318 & 205 & 235 & 268 & -- \\
\bottomrule
\end{tabular}
\end{table}
\FloatBarrier

\subsection{Clipping decides the per-tensor 4-bit result on Maze}
\label{app:depth-ingolfsson}

Concurrent work reports that per-tensor W4A4 collapses Sudoku but not Maze \citep{ingolfsson2026quantizing}. With
activation scales clipped at the maximum over calibration calls, our Maze model scores 0.0\%; with a 99.99th-percentile
scale it scores 82.2\%, and which layers are quantized moves accuracy by at most 0.1 points (Table~\ref{tab:app-clip}).
Under percentile clipping, the Maze gap to full precision closes from $-40.6$ to $-6.6$ points over the first
supervision steps and then holds at $-6.2$ and $-6.4$: the improvement with depth that the concurrent work reports is
this early closing of a constant gap.

\begin{table}[!htbp]
\caption{\textbf{The clipping rule alone decides whether per-tensor W4A4 survives on Maze.} Exact \% at 16
supervision steps, mean [min--max] over 3 calibration draws, each with its own full-precision reference;
$\rho$ = per-call output error. Maze-attention $n{=}500$ (full precision 88.4), Sudoku-MLP $n{=}1000$ (full
precision 71.4); A100 platform.}
\label{tab:app-clip}
\centering\scriptsize
\begin{tabular}{@{}llllll@{}}
\toprule
Clipping & Quantized layers & Maze exact & Maze valid path & Sudoku-MLP exact & $\rho$ Maze / MLP \\
\midrule
max over calls (ours) & all linears & 0.0 [0.0--0.0] & 0.0 & 0.0 [0.0--0.0] & 0.293 / 0.466 \\
max over calls & block only & 0.0 [0.0--0.0] & 0.0 & 0.0 [0.0--0.0] & 0.293 / 0.465 \\
99.99th percentile & all linears & 82.2 [82.2--82.2] & 98.4 & 0.8 [0.4--1.0] & 0.145 / 0.270 \\
99.99th percentile & block only & 82.3 [82.2--82.4] & 98.4 & 0.8 [0.5--1.0] & 0.144 / 0.269 \\
\bottomrule
\end{tabular}
\end{table}
\FloatBarrier

\section{Fixed versus fresh error}
\label{app:fixedfresh}

Figure~\ref{fig:fixedfresh}a plots the Sudoku-MLP rows of Table~\ref{tab:app-arms-mlp}, and
Figure~\ref{fig:fixedfresh}b the fidelity-matched rows of Table~\ref{tab:app-fixedfresh}. 

\subsection{More loops repair a fresh error and leave a fixed one, except on some attention draws}
\label{app:fixedfresh-noise}

We give three Sudoku models a fixed error (weight noise) and a fresh error (activation noise) at levels whose fidelity
at 512 supervision steps overlaps, and follow the gap to 512 steps (Table~\ref{tab:app-fixedfresh}). At the two smaller
fresh-noise levels of each model, all 30 draws close part of their gap, to within two points at the smallest level of
the MLP and HRM models; fresh noise stops catching up once fidelity falls to about 0.5--0.7. Fixed noise at matched
fidelity closes no gap on the MLP mixer (0 of 15 draws) or on HRM (0 of 15) and closes part of it on 5 of 15 attention
draws. Rounding behaves like a fixed error. Among the 43 quantized models of the A100 deep set, 7 track full precision,
16 keep a constant gap, 14 widen, and the only 6 that catch up are the three surviving 3-bit attention formats (fully)
and three partial cases; no quantized MLP or HRM model catches up. 
\label{app:fixedfresh-attn}

\begin{table}[!htbp]
\caption{\textbf{Fresh noise catches up with more loops; fixed noise does not, except on some attention
draws.} Gap to full precision (points) at 16 / 64 / 512 supervision steps and $D = $ gap(512) $-$ gap(16),
mean over 5 noise draws [95\% $t$-interval across draws]; fidelity $\varphi_H$ at 512 steps in parentheses;
per-draw regimes by the rule of Appendix~\ref{app:protocol-stats}. 1800 stratified Sudoku-Extreme puzzles; L40S
platform; each model with its own full-precision reference.}
\label{tab:app-fixedfresh}
\centering\scriptsize
\begin{tabular}{@{}lllllp{0.22\textwidth}@{}}
\toprule
Model & Error & $\sigma$ ($\varphi_H$) & Gap 16 / 64 / 512 & $D$ & Per-draw regimes \\
\midrule
TRM MLP & fresh & 0.15 (0.89) & $-4.4$ / $-1.4$ / $-0.1$ & $+4.3$ [$+3.8$, $+4.7$] & full catch-up 5 \\
 & fresh & 0.2 (0.87) & $-7.8$ / $-3.6$ / $-1.4$ & $+6.4$ [$+5.6$, $+7.2$] & full 1, partial 4 \\
 & fresh & 0.25 (0.70) & $-17.1$ / $-17.1$ / $-18.4$ & $-1.3$ [$-1.9$, $-0.8$] & constant 5 \\
 & fixed & 0.08 (0.88) & $-3.4$ / $-2.8$ / $-3.6$ & $-0.2$ [$-0.8$, $+0.3$] & constant 5 \\
 & fixed & 0.1 (0.85) & $-6.1$ / $-5.6$ / $-6.6$ & $-0.4$ [$-1.4$, $+0.5$] & constant 5 \\
 & fixed & 0.12 (0.72) & $-17.8$ / $-19.1$ / $-22.0$ & $-4.2$ [$-6.5$, $-1.9$] & widening 5 \\
\midrule
TRM attention & fresh & 0.4 (0.89) & $-8.5$ / $-5.6$ / $-2.3$ & $+6.1$ [$+5.3$, $+7.0$] & partial catch-up 5 \\
 & fresh & 0.5 (0.81) & $-25.1$ / $-21.4$ / $-16.4$ & $+8.7$ [$+8.1$, $+9.3$] & partial catch-up 5 \\
 & fresh & 0.6 (0.51) & $-72.2$ / $-78.8$ / $-84.1$ & $-11.9$ [$-12.4$, $-11.4$] & widening 5 \\
 & fixed & 0.5 (0.86) & $-7.9$ / $-7.1$ / $-5.3$ & $+2.6$ [$-1.8$, $+7.1$] & partial 2, constant 3 \\
 & fixed & 0.6 (0.80) & $-16.1$ / $-15.5$ / $-13.2$ & $+2.9$ [$-2.5$, $+8.3$] & full 1, partial 1, constant 3 \\
 & fixed & 0.7 (0.71) & $-31.3$ / $-31.5$ / $-29.8$ & $+1.5$ [$-11.0$, $+14.0$] & partial 1, constant 3, widening 1 \\
\midrule
HRM & fresh & 0.4 (0.90) & $-17.2$ / $-9.5$ / $-1.4$ & $+15.8$ [$+15.1$, $+16.5$] & full catch-up 5 \\
 & fresh & 0.5 (0.76) & $-43.9$ / $-40.9$ / $-40.8$ & $+3.2$ [$+2.6$, $+3.7$] & partial catch-up 5 \\
 & fresh & 0.6 (0.59) & $-53.4$ / $-59.2$ / $-64.8$ & $-11.4$ [$-11.6$, $-11.2$] & widening 5 \\
 & fixed & 0.5 (0.91) & $-7.2$ / $-7.5$ / $-9.3$ & $-2.1$ [$-3.5$, $-0.7$] & widening 3, constant 2 \\
 & fixed & 0.7 (0.86) & $-15.1$ / $-16.9$ / $-19.2$ & $-4.1$ [$-5.9$, $-2.2$] & widening 4, constant 1 \\
 & fixed & 0.85 (0.81) & $-23.7$ / $-26.1$ / $-28.9$ & $-5.2$ [$-6.8$, $-3.6$] & widening 5 \\
\bottomrule
\end{tabular}
\end{table}
\FloatBarrier

\subsection{Fresh zero-mean noise alone produces the cliff; a bias or a fixed draw only moves it}
\label{app:fixedfresh-arms}

Concurrent work attributes the 4-bit collapse to a bias that accumulates with reuse \citep{ingolfsson2026quantizing}.
Fresh zero-mean noise has no bias, yet it takes Sudoku-MLP from 62.9\% to 2.3\% between $\sigma$ 0.2 and 0.5 with a
draw spread of at most 2 points, and Maze from 85.5\% at 0.7 to 23.3\% at 1.0 (Tables~\ref{tab:app-arms-mlp}
and~\ref{tab:app-arms-maze}). Keeping the activation noise fixed across calls moves Sudoku-MLP's half-accuracy point
from 0.29 to 0.22, and a sign-consistent bias of equal norm moves it to 0.15--0.19. Measured by the per-call output
error $\rho$, the Sudoku-MLP variants collapse within a band three times wide (0.038--0.124), against eight times wide in
$\sigma$; per unit $\rho$ the bias is no more harmful than fresh noise. Fresh zero-mean noise also collapses
Sudoku-attention (44.0\% at 0.5 to 1.1\% at 0.7) and HRM (12.0\% to 0\%), and on HRM a fixed zero-mean draw and a bias
of equal norm are indistinguishable ($\sigma_{1/2}$ 0.24--0.35 and 0.25--0.36). The extra damage of the fixed variants
comes from their being fixed across loops, not from a nonzero mean. 

\begin{table}[!htbp]
\caption{\textbf{TRM Sudoku-MLP: zero-mean fresh noise alone produces the cliff; bias and fixedness move it.}
Exact \% at each $\sigma$, mean [min--max] over 3 noise draws; $\sigma_{1/2}$ and $\rho_{1/2}$ are where exact
accuracy falls below half of full precision (71.4), per draw. Weight bias (\texttt{wb}, own grid): $\sigma$ 0.02
$\to$ 66.3 [64--69], 0.05 $\to$ 3.1 [2--4]; $\sigma_{1/2}$ 0.034--0.035, $\rho_{1/2}$ 0.047--0.049. $n{=}1000$,
$H{=}3$, $L{=}6$, 16 supervision steps; A100 platform.}
\label{tab:app-arms-mlp}
\centering\scriptsize
\setlength{\tabcolsep}{3pt}
\begin{tabular}{@{}llllllll@{}}
\toprule
Variant & $\sigma$ 0.1 & 0.15 & 0.2 & 0.3 & 0.5 & $\sigma_{1/2}$ & $\rho_{1/2}$ \\
\midrule
\texttt{an} fresh, zero-mean & 70.1 [70--70] & 67.4 [67--68] & 62.9 [63--63] & 32.9 [32--34] & 2.3 [2--2] & 0.287--0.293 & 0.080--0.082 \\
\texttt{af} fixed, zero-mean & 66.5 [66--67] & 58.6 [57--60] & 43.8 [41--46] & 10.4 [8--13] & 0.3 & 0.216--0.231 & 0.061--0.064 \\
\texttt{ab} fixed, sign-consistent & 64.6 [62--67] & 46.5 [35--56] & 22.9 [10--31] & 3.7 [0--6] & 0.0 & 0.149--0.190 & 0.094--0.124 \\
\texttt{wn} weight, zero-mean & 64.9 [63--66] & 26.5 [24--31] & 5.5 [4--7] & 1.0 & 0.0 & 0.135--0.144 & 0.038--0.042 \\
\bottomrule
\end{tabular}
\end{table}
\begin{table}[!htbp]
\caption{\textbf{TRM Maze-attention: the same variants, a wider tolerance.} Exact \% mean [min--max] over 3 noise
draws; $\sigma_{1/2}$ on exact match and on valid path; full precision 88.4. $n{=}500$, $H{=}3$, $L{=}4$; A100
platform.}
\label{tab:app-arms-maze}
\centering\scriptsize
\setlength{\tabcolsep}{3pt}
\begin{tabular}{@{}llllllll@{}}
\toprule
Variant & $\sigma$ 0.5 & 0.7 & 1.0 & 1.5 & $\sigma_{1/2}$ exact & $\sigma_{1/2}$ valid path & $\rho_{1/2}$ \\
\midrule
\texttt{an} fresh, zero-mean & 87.5 [87--88] & 85.5 [85--87] & 23.3 [23--24] & 0.0 & 0.897--0.902 & 1.20--1.21 & 0.198--0.199 \\
\texttt{af} fixed, zero-mean & 77.7 [70--82] & 34.9 [20--48] & 0.0 & 0.0 & 0.60--0.73 & 0.78--0.83 & 0.13--0.16 \\
\texttt{ab} fixed, sign-consistent & 86.1 [84--88] & 79.5 [72--84] & 9.3 [0--25] & 0.0 & 0.82--0.90 & 0.83--0.91 & 0.18--0.20 \\
\texttt{wn} weight, zero-mean & 87.7 [87--88] & -- & 86.1 [85--87] & 27.1 [0--81] & 1.24--1.73 & 1.73--1.75 & 0.30--0.44 \\
\bottomrule
\end{tabular}
\end{table}
\FloatBarrier

\subsection{Attention does not become one-hot, so one-hot maps do not explain the exception}
\label{app:fixedfresh-onehot}

Attention maps that sharpen to one position at the settled state would make their rounding error vanish as the model
settles, which would explain why some fixed errors on the attention mixer close their gap. We wrote the prediction and
its pass criterion before any record existed. The criterion was that in each catch-up model the error of the attention
path at step 512 falls below half its step-1 value and the attention entropy falls, and that in constant-gap and MLP
models it stays at least 0.8 times its step-1 value. Neither holds (Table~\ref{tab:app-onehot}). From step 4 to 512 the normalised entropy stays between
0.62 and 0.68 and the mean maximum probability near 0.26, at full precision and in every compressed model; the
attention-path error rises from step 1 to 4 and is then flat, with the same shape in catch-up and constant-gap models;
the whole-step error roughly halves in every model alike.

\begin{table}[!htbp]
\caption{\textbf{Attention never becomes one-hot, and its error does not shrink in the models that catch up.}
Values at supervision steps 1 $\to$ 512 (mixer-path error also at 4 / 16 / 64). $n{=}128$ Sudoku-Extreme puzzles,
512 supervision steps; L40S platform; hypothesis and pass criterion written before any record they score existed.}
\label{tab:app-onehot}
\centering\scriptsize
\setlength{\tabcolsep}{2.5pt}
\begin{tabular}{@{}lllll@{}}
\toprule
Model / format (regime) & Mixer-path error, steps 1 / 4 / 16 / 64 / 512 & Whole-step error & Entropy & Top probability \\
\midrule
attention, full precision & 0 & 0 & 0.694 $\to$ 0.643 & 0.21 $\to$ 0.26 \\
attention w3c (catch-up) & 0.210 / 0.281 / 0.287 / 0.290 / 0.291 & 0.636 $\to$ 0.289 & 0.719 $\to$ 0.646 & 0.19 $\to$ 0.26 \\
attention w3a (catch-up) & 0.170 / 0.232 / 0.237 / 0.239 / 0.240 & 0.444 $\to$ 0.229 & 0.703 $\to$ 0.643 & 0.20 $\to$ 0.26 \\
attention w3g32 (catch-up) & 0.136 / 0.181 / 0.187 / 0.189 / 0.190 & 0.310 $\to$ 0.168 & 0.696 $\to$ 0.638 & 0.21 $\to$ 0.26 \\
attention w4t (constant) & 0.169 $\to$ 0.240 & 0.503 $\to$ 0.229 & 0.711 $\to$ 0.651 & 0.20 $\to$ 0.25 \\
MLP w4c (constant) & 0.086, flat & 0.520 $\to$ 0.220 & -- & -- \\
MLP w4a (constant) & 0.076, flat & 0.437 $\to$ 0.181 & -- & -- \\
MLP w4g32 (constant) & 0.046, flat & 0.382 $\to$ 0.128 & -- & -- \\
\bottomrule
\end{tabular}
\end{table}
\FloatBarrier


\section{Settling and return}\label{app:return}

Throughout, $f_\theta$ is the full-precision loop (fp32; bf16 for the language models), $f_{\tilde\theta}$ the
compressed one, and $z^\star$ and $\tilde z^\star$ their settled states after the same loops on the same input.
\emph{Retained accuracy} is compressed accuracy over full-precision accuracy on the same examples; a compressed model
collapses when it retains less than 0.5. Models whose full-precision accuracy is below 10\% are excluded. The L40S and
A100 platforms are run and scored separately, and no number in this appendix pools them.

\subsection{A collapsed model still settles}\label{app:return_contraction}

Two explanations of why a model collapses while its state stays within reach were written, with their bars, before any
record they score existed: (A) \emph{contraction loss}, the compressed loop stops settling; (B) \emph{room}, the loop
still settles and its shift carries the state past a model-specific room. We measure the late contraction ratio $r$,
the median over the last three loops of $\lVert z_{t+1}-z_t\rVert/\lVert z_t-z_{t-1}\rVert$ on the compressed
trajectory, on every example full precision solves. Only 12\% of collapsed models have $r\ge1$, against the 80\% that
(A) needs, and 88\% keep settling, nearly the rate of surviving ones (91\%), so (A) fails and (B)'s first clause holds
(Table~\ref{tab:app_ab}). (B)'s zero-parameter width, a room read off at $\sigma_0{=}0.05$, misses the new seeds'
tolerance by a factor of 8.9, because the settling shift grows faster than linearly with the noise level. The implicit
models settle as well: Broyden reaches its residual tolerance on nearly every image at every noise level, collapsed
models included. 
Several full-precision models are not fully settled at 16 loops (late ratio of the full-precision trajectory
0.93--1.00 for TRM Sudoku MLP/attention, DeepThinking and one LSQ model), so $r\approx1$ alone does not show that the
loop stopped settling there. 

\begin{table}[H]
\caption{\textbf{Collapsed models keep settling: the contraction-loss explanation fails its defining clause.}
Each row is a clause with its bar, the outcome and the verdict. Width errors are on the six new mixer seeds
(\S\ref{app:widthlaw_seeds}; median $|\log|$ ratio, as a factor); bias = predicted/measured tolerance, MLP / attention.
L40S platform, 32 record files, 201 unique compressed models (60 collapsed).} 
\label{tab:app_ab}
\centering\footnotesize
\begin{tabular}{@{}llll@{}}
\toprule
Clause & Bar & Outcome & Verdict \\
\midrule
(A1) collapsed models with $r\ge1$ & $\ge$80\% & 12\% & fails \\ 
(A2) surviving models with $r<1$ & $\ge$80\% & 91\% & holds \\ 
(B1) collapsed models with $r<1$ & $\ge$80\% & 88\% & holds \\ 
(B2) zero-parameter width, new seeds & bias in [0.9, 1.1] & $\times$8.9 (bias 8.25 / 9.60) & fails \\ 
\quad one fitted scale & (reported) & $\times$1.57 (bias 1.45 / 1.68) & -- \\ 
\quad reference: $\delta^\star/S$, $\delta^\star/S_{\text{mix}}$ & -- & $\times$1.39, $\times$1.16 & -- \\ 
\bottomrule
\end{tabular}
\end{table}
\subsection{Predictions written before their records}\label{app:return_prereg}

The return test starts the full-precision loop from $z(t)=z^\star+t\,(\tilde z^\star-z^\star)$ for $t$ on a grid up to
16 and runs $k{=}16$ full-precision loops (DeepThinking: its own depth; MDEQ: a Broyden budget). An example
\emph{returns} at $t$ if the decoded answer equals the full-precision answer from $z^\star$ after the same loops; $t_c$
is the largest grid $t$ up to which every grid point returns, $\rho=1/t_c$, and a compressed model's $\rho$ is the
median over the examples full precision solves. Table~\ref{tab:app_prereg} lists every prediction with its scoring rule
and outcome, the falsified ones included. P1 fails because collapsed states return. On the L40S platform 29 of 64
collapsed models return even from the grid cap, as a cross-seeding record disclosed before P1 was scored (the
full-precision loop run from the perturbed end state recovered 95--99\% of the Sudoku puzzles full precision solves).
The per-example link is one-sided, with $P(\text{correct}\mid\text{no return})$ 0.00--0.14 and
$P(\text{correct}\mid\text{return})$ 0.62--0.84. For P3, the halting logits saturate in 78\% (L40S) and 80\% (A100) of
MLP examples and 23\% and 28\% of HRM examples, so the $t_c$ comparison rests on few examples there; the correct-draw
result is unaffected. 

\begin{table}[H]
\caption{\textbf{Every prediction and its outcome, the falsified ones included.} Order is the order in which the
predictions and their scoring rules were written, each before any record it scores existed. P1--P4 are scored on the
L40S return records (255 compressed models; 191 scored, 64 collapsed); the A100 replicate covers the released models
(148; 91 scored, 28 collapsed) and scores P1 and the head-pick part of P3 only. The two platforms are scored
separately. The looped-LM return test (\S\ref{app:lms_return}) has its own predictions, also written before any record
they score existed.} 
\label{tab:app_prereg}
\centering\footnotesize
\begin{tabular}{@{}p{0.05\linewidth}p{0.08\linewidth}p{0.33\linewidth}p{0.42\linewidth}@{}}
\toprule
ID & Order & Prediction and bar & Outcome \\
\midrule
P1 & 1 & A compressed model collapses iff its settled state does not return under the full-precision loop: rule
``collapse iff median $\rho>1$'' balanced accuracy $\ge0.80$; AUROC($\rho$) $>$ AUROC($-\varphi$); leave-one-family-out
$\rho\ge\varphi$; Spearman(retained, return) $\ge0.7$. &
\textbf{Falsified} on both platforms. L40S: balanced accuracy 0.63; AUROC $\rho$ 0.70 vs $-\varphi$ 0.98; LOFO 0.72 vs 0.94; 29/64 collapsed models return from the grid cap. A100: 0.61; AUROC 0.60 vs 0.97; LOFO 0.78 vs 0.91; 17/28 at the cap. \\ 
P2 & 1 & Finishing law: gain of full-precision finishing = return rate $\times$ gap; supported if $|$pred$-$obs$|\le\max(2\text{ pts},25\%)$ on $\ge$80\% of matched models. &
\textbf{Confirmed}: 20/20 matched models, mean $|$pred$-$obs$|$ 1.52 points, $r=0.997$ (L40S). \\ 
P3 & 1 & Sampling: the halting head picks draws with larger $t_c$ (sign test $p<0.05$ in each model), and the best-of-16 gain peaks at intermediate return rates. &
\textbf{Falsified}. The head picks larger-$t_c$ draws in HRM w3c (57/18) and MLP w4c (15/1), not attention w3c (64/67); the gain is largest at the lowest return rate. The head does pick correct draws (picked-correct vs median draw: attention 68/0, HRM 89/0, MLP 62/0 on the L40S platform; 72/0, 83/1, 68/1 on the A100 platform). \\ 
P4 & 1 & Radius law: $\sigma_{1/2}\approx0.1\times$ median $t_c$ at noise 0.1 removes the tolerance law's mixer bias. &
\textbf{Falsified}: $R^2(\log)=-0.84$, worse than the laws with $S$ and $S_{\text{mix}}$. \\ 
K1 & 2 & $k\le4$ full-precision finishing steps recover $\ge$80\% of the examples full precision solves, on collapsed models. &
\textbf{Falsified} on the L40S platform, \textbf{partly} on the A100 platform; 8--16 steps needed (Table~\ref{tab:app_k_babel}). \\ 
K2 & 2 & w8c finishing within 2 points of full-precision finishing on $\ge$90\% of compressed models (at $k=4$ and 16). &
\textbf{Partly} on the L40S platform (0.84 of models), \textbf{confirmed} on the A100 platform (0.93); median $|$w8c$-$fp32$|$ 0.003 (L40S), 0.000 (A100). \\ 
A / B & 3 & Contraction loss vs room (Table~\ref{tab:app_ab}). & A rejected; B clause 1 holds, clause 2 fails. \\ 
\bottomrule
\end{tabular}
\end{table}
\subsection{Recovery climbs with the finishing steps, and 8-bit finishing recovers as much as full precision}\label{app:return_k}

From $\tilde z^\star$ we run the full-precision, w8c or w4c loop for $k\in\{1,2,4,8,16\}$ finishing steps. Recovery
needs 8--16 steps against the 1--4 of prediction K1, and the two platforms agree (Table~\ref{tab:app_k_babel}). The
families differ (L40S recovery at $k{=}2/4/16$): TRM-attention 0.92/0.93/0.93, where two steps suffice; TRM-MLP
0.58/0.77/0.90; trained seeds 0.58/0.73/0.94; HRM 0.31/0.54/0.82; quantization-aware-trained models 0.31/0.52/0.73;
MDEQ 0.32/0.76/1.00; DeepThinking 0/0/0.04, since it runs 100 loops at test time and its return test at $k{=}100$
returns. After 16 steps, w8c finishing retains 0.95 against 0.96 for full precision on both platforms, and w4c retains
0.90 (L40S) and 0.87 (A100) (Figure~\ref{fig:return}a). The K2 bar, within 2 points of full-precision finishing on 90\%
of models, is met on the A100 platform (0.93) and narrowly missed on the L40S platform (0.84), where the misses are 2--4
points. The controller repeats the comparison at deployment scale (Appendix~\ref{app:controller-tab2}). 
\label{app:return_w8}

\begin{table}[H]
\caption{\textbf{Recovery on collapsed models climbs with the number of finishing steps, on both platforms alike.}
Medians over collapsed models, full-precision finishing; \emph{recovery} = share of the examples full precision solves
whose decoded answer equals full precision's after the same $k$ loops; \emph{retained} = accuracy after finishing over
full-precision accuracy. The compressed model alone retains 0.05 (L40S) and 0.01 (A100); surviving models retain 0.97
and 1.00 already at $k{=}1$. L40S platform: 36 record files, 201 compressed models with full-precision accuracy
$\ge$10\%, 65 collapsed (TRM released checkpoints, trained width and seed variants, quantization-aware-trained and
fine-tuned models, HRM, on Sudoku-Extreme and Maze-Hard; DeepThinking mazes; MDEQ-Small/XL). A100 platform: 12 files,
100 compressed models, 28 collapsed (released models). The two platforms are never pooled.} 
\label{tab:app_k_babel}
\centering\footnotesize
\begin{tabular}{@{}llccccc@{}}
\toprule
Platform & $k$ (full-precision finishing steps) & 1 & 2 & 4 & 8 & 16 \\
\midrule
L40S & recovery (median model) & 0.28 & 0.50 & 0.71 & 0.83 & 0.89 \\ 
     & retained accuracy       & 0.29 & 0.51 & 0.75 & 0.88 & 0.96 \\ 
\midrule
A100 & recovery (median model) & 0.32 & 0.67 & 0.80 & 0.87 & 0.88 \\ 
     & retained accuracy       & 0.33 & 0.69 & 0.81 & 0.88 & 0.96 \\ 
\bottomrule
\end{tabular}
\end{table}
\subsection{The finishing law predicts the gain on every model it can score}\label{app:return_law}

Finishing repairs an example when the compressed model gets it wrong, full precision gets it right, and the compressed
settled state returns, so the expected gain of a compressed model is
\[
\widehat{\Delta} \;=\; \frac{1}{n}\sum_{i=1}^{n}\mathrm{ret}_i\,(c^{f}_i - c^{q}_i),
\]
with $\mathrm{ret}_i$ the return indicator at $t{=}1$ and $c^f_i,c^q_i$ full-precision and compressed correctness. The
observed gain comes from an independent run, 16 compressed loops followed by 8 full-precision finishing steps against 16
compressed loops, on the first 256 puzzles; a model is \emph{matched} when that run's compressed per-example correctness
equals the return run's on at least 98\% of examples, and the rule fixed in advance the difference between the return
test's 16 loops and the finishing run's 8. All 20 matched TRM and HRM models fall within the tolerance of
max(2 points, 25\%), with mean absolute error 1.52 points and $r{=}0.997$ (L40S platform; Figure~\ref{fig:return}b).
Where the state does not return, the gain vanishes: HRM Sudoku w3t has return 0.00 and gain 0.000, and TRM-attention w3t
return 0.25 and gain 0.20 of a 0.73 gap. For MDEQ-XL w4g32, whose settling shift is tiny, the law predicts 0.008 and the
run gives 0.006 (unmatched at 64\% agreement; reported, not scored). 
\FloatBarrier

\section{The tolerance law}\label{app:widthlaw}
\label{app:widthlaw_defs}

Relative weight noise at level $\sigma$ adds Gaussian noise with standard deviation $\sigma$ times each tensor's own
standard deviation, drawn once and kept across loops. The tolerance $\sigma_{1/2}$ is the level at which accuracy falls
to half of full precision (three draws, median per level; for the DEQ-Transformer, the level at which perplexity
doubles). The push $\delta$ is the relative error of one noisy loop at the full-precision settled state (Eq.~\ref{eq:push}),
and the sensitivity is $S=\mathrm{median}\,\delta/\sigma_0$ at $\sigma_0{=}0.05$ (median over three draws of the median
over inputs; $n{=}256$). The law is $\sigma_{1/2}=\delta^\star/S$ with one shared room $\delta^\star$. Each model's
$\sigma_{1/2}$ and $S$ come from one platform; $S$ measured on both platforms agrees within 2.7\% for 18 models, and the
push of each compressed model within 0.05\% (median) and 3.3\% (max) over 279 compressed models. 

\subsection{The implied room spans 0.17 to 0.89 across the 19 model--task pairs}\label{app:widthlaw_all}

Tables~\ref{tab:app_wl_babel} and~\ref{tab:app_wl_p4d} list every model--task pair in the fit, one table per platform,
with its implied room $S\cdot\sigma_{1/2}$, the push at which its accuracy halves. Among the TRMs, the MLP-mixer models
sit at the low end (0.17--0.37) and the attention models higher (0.51--0.55), the mixer effect that
\S\ref{app:widthlaw_seeds} tests on new seeds. 

\begin{table}[H]
\caption{\textbf{L40S models: the implied room spans 0.19 to 0.89.} full precision = full-precision accuracy (DEQ-Transformer:
perplexity, lower is better); $\sigma_{1/2}$ = measured tolerance; $S(0.05)$ = sensitivity at $\sigma_0=0.05$;
implied room = $S\cdot\sigma_{1/2}$. L40S platform.} 
\label{tab:app_wl_babel}
\centering\footnotesize
\begin{tabular}{@{}llcccc@{}}
\toprule
Model & Family & Full precision & $\sigma_{1/2}$ & $S(0.05)$ & $S\cdot\sigma_{1/2}$ \\
\midrule
TRM Sudoku MLP & TRM & 0.715 & 0.14 & 1.74 & 0.24 \\
TRM Sudoku attention & TRM & 0.727 & 0.80 & 0.69 & 0.55 \\
TRM Maze attention & TRM & 0.875 & 1.71 & 0.32 & 0.55 \\
HRM Sudoku & HRM & 0.477 & 1.05 & 0.35 & 0.37 \\
HRM Maze & HRM & 0.748 & 1.71 & 0.23 & 0.40 \\
DeepThinking maze 13 & DeepThinking & 1.000 & 0.37 & 0.92 & 0.34 \\
DeepThinking maze 33 & DeepThinking & 1.000 & 0.26 & 0.93 & 0.25 \\
MDEQ-Small & DEQ & 0.738 & 0.23 & 3.06 & 0.70 \\
MDEQ-XL & DEQ & 0.755 & 0.37 & 2.41 & 0.89 \\
DEQ-Transformer & DEQ & 23.1 (perplexity) & 0.34 & 0.55 & 0.19 \\
\bottomrule
\end{tabular}
\end{table}
\begin{table}[H]
\caption{\textbf{A100 models: the nine scored Sudoku variants, with implied room 0.17 to 0.52.} Columns as in
Table~\ref{tab:app_wl_babel}. Excluded from the fit: attention w256 seeds 0--2 (full-precision accuracy 10--28\%,
undertrained) and attention w512 s0 ($\sigma_{1/2}>0.7$, not located). A100 platform, $\sigma_{1/2}$ from $n=1000$.} 
\label{tab:app_wl_p4d}
\centering\footnotesize
\begin{tabular}{@{}lcccc@{}}
\toprule
Variant & Full precision & $\sigma_{1/2}$ & $S(0.05)$ & $S\cdot\sigma_{1/2}$ \\
\midrule
attention w512 s1 & 0.629 & 0.65 & 0.80 & 0.52 \\
attention w512 s2 & 0.534 & 0.60 & 0.85 & 0.51 \\
MLP w256 s0 & 0.599 & 0.14 & 1.46 & 0.21 \\
MLP w256 s1 & 0.623 & 0.13 & 2.05 & 0.27 \\
MLP w256 s2 & 0.640 & 0.13 & 1.30 & 0.17 \\
MLP w512 s0 & 0.836 & 0.16 & 1.50 & 0.25 \\
MLP w512 s1 & 0.825 & 0.16 & 1.96 & 0.32 \\
MLP w512 s2 & 0.842 & 0.13 & 1.60 & 0.20 \\
MLP w768 s0 & 0.753 & 0.15 & 2.47 & 0.37 \\
\bottomrule
\end{tabular}
\end{table}
\subsection{One shared room predicts tolerance to a median factor of 1.5}\label{app:widthlaw_fits}

Over the 19 model--task pairs, $\delta^\star=0.344$, $R^2(\log$, slope $-1)=0.72$, and a free fit gives
$\log\sigma_{1/2}=-1.07-1.03\log S$; the leave-one-model-out error is $\times1.52$ at the median and $\times2.74$ at
worst (MDEQ-XL). On one platform only, the L40S fit (10 pairs) gives $\delta^\star=0.40$, $R^2=0.67$, free slope
$-0.83$, leave-one-out $\times1.57$ / $\times2.45$, and the A100 fit (9 models) $\delta^\star=0.29$, $R^2=0.62$, free
slope $-1.42$, $\times1.46$ / $\times1.93$. Held out by family and fit on the rest, the median error is HRM $\times1.13$
(worst $\times1.18$), DeepThinking $\times1.22$ ($\times1.43$), TRM $\times1.42$ ($\times2.37$) and DEQ $\times2.18$
($\times2.78$); a fit on TRM and HRM alone predicts DeepThinking and DEQ to $\times1.72$ ($\times2.74$). Within a family
the law therefore predicts tolerance to within about $\times1.1$--$1.4$; across families the room differs, with the
implicit models tolerating more than their sensitivity predicts (MDEQ by about $\times2$--$3$) and the DEQ-Transformer
less. As a collapse test on quantized models, a model is predicted to collapse when its measured push exceeds the room
fitted on noise tolerances. On the A100 platform TRM (91 compressed models) scores accuracy 0.89 and AUROC 0.98 and HRM
(26) 0.92 and 1.00; on the L40S platform DeepThinking (8) scores 1.00 and DEQ (18) 0.72 with AUROC 0.82.

\subsection{Parameter-free predictions for new mixer seeds land inside the error band}\label{app:widthlaw_seeds}

Six TRM Sudoku seeds at width 512, three per mixer, were trained after the prediction rule
$\sigma_{1/2}=0.4033/S(0.05)$ was fixed, with $\delta^\star$ from the five L40S TRM/HRM models and an error band from
their leave-one-out errors ($\times1.46$ median, $\times1.92$ worst); the pipeline wrote each prediction from the seed's
own $S$ before our noise sweep of that seed. The rule has no free parameter, and the predictions are parameter-free
and not blind, since five of the six seed evaluation files already existed, unread, when they were written. All six
land inside the $\times1.92$ band and three inside $\times1.46$, with the order right and the error signed by mixer,
MLP predicted too wide and attention too narrow, as in the fit set (Table~\ref{tab:app_seeds}). Attention tolerates
about 4.1$\times$ the weight noise of the MLP mixer on the L40S platform; on the A100 platform the w512 seeds 0--2 show
the same order ($\sigma_{1/2}$ attention $>0.7$/0.65/0.60, MLP 0.16/0.16/0.13), per-tensor INT4 collapses every MLP
seed on both platforms (A100 w4t 3.0/2.6/1.2), and the sensitivity differs by mixer ($S(0.1)$ attention 0.68--0.76,
MLP 1.59--1.83). 

\begin{table}[H]
\caption{\textbf{New mixer seeds: attention tolerates about 4$\times$ the weight noise of the MLP mixer, and every
parameter-free prediction lands inside the band.} Full-precision and w4t (per-tensor INT4) accuracy from the seed evaluation,
$\varphi$ in parentheses; measured $\sigma_{1/2}$ from our noise sweep; error = measured/predicted as a factor. L40S
platform, $n=1000$, three noise draws.} 
\label{tab:app_seeds}
\centering\footnotesize
\begin{tabular}{@{}lccccc@{}}
\toprule
Seed & Full precision & w4t ($\varphi$) & predicted $\sigma_{1/2}$ & measured $\sigma_{1/2}$ & error \\
\midrule
attention s3 & 68.6 & 27.9 (0.741) & 0.537 & 0.709 & $\times$1.32 \\
attention s4 & 71.1 & 60.1 (0.840) & 0.563 & 0.578 & $\times$1.03 \\
attention s5 & 64.3 & 57.7 (0.901) & 0.603 & 0.723 & $\times$1.20 \\
MLP s3 & 82.1 & 3.9 (0.559) & 0.268 & 0.176 & $\times$1.52 \\
MLP s4 & 81.3 & 6.1 (0.534) & 0.284 & 0.158 & $\times$1.80 \\
MLP s5 & 82.4 & 7.8 (0.498) & 0.236 & 0.147 & $\times$1.60 \\
\bottomrule
\end{tabular}
\end{table}
\subsection{The token mixer's sensitivity shrinks the mixer bias but keeps its sign (post hoc)}\label{app:widthlaw_smix}

The TRM MLP model's fragility sits in its token mixer, since noise on the token-mixing MLP alone collapses it. A
sensitivity measured with noise on the token mixer only, $S_{\text{mix}}$, raises $R^2$ from 0.84 to 0.91 on the 15
L40S fit models and shrinks the mixer bias by about 40\% in log terms (45\% for MLP, 39\% for attention) while keeping
its sign (Table~\ref{tab:app_smix}). We chose this law after the bias was known, so its gains are post hoc; no
parameter used the new seeds. $S_{\text{mix}}/S$ itself depends on the mixer (MLP 0.71--0.98, attention 0.48--0.72, HRM
0.67--0.72), so the mixer's share of the sensitivity differs by architecture. 

\begin{table}[H]
\caption{\textbf{The token mixer's sensitivity shrinks the mixer bias but keeps its sign (post hoc).} LOO = leave one
model out; errors are factors; bias = predicted/measured tolerance on the fit set, where above 1 over-predicts. L40S
platform only; 15 fit models (5 released TRM/HRM and 10 variants with full-precision accuracy $\ge$30\%); the six new
seeds held out.} 
\label{tab:app_smix}
\centering\footnotesize
\begin{tabular}{@{}lccccc@{}}
\toprule
Law & $R^2$ & LOO median / worst & new seeds median / worst & bias MLP / attn / HRM \\
\midrule
$\sigma_{1/2}=\delta^\star/S$ & 0.84 & $\times$1.47 / $\times$2.07 & $\times$1.39 / $\times$1.61 & 1.37 / 0.64 / 0.89 \\
$\sigma_{1/2}=\delta^\star/S_{\text{mix}}$ & 0.91 & $\times$1.35 / $\times$1.96 & $\times$1.16 / $\times$1.29 & 1.19 / 0.76 / 0.98 \\
\bottomrule
\end{tabular}
\end{table}
\subsection{LSQ training lowers the sensitivity and leaves the room unchanged}\label{app:widthlaw_lsq}

Quantization-aware training with learned step sizes \citep{esser2020lsq} raises the full-precision model's tolerance
from 0.135 to about 0.23. The law allows two routes, a lower sensitivity or more room, and the measurement picks one:
$S$ falls by 1.71$\times$, 1.30$\times$ and 1.65$\times$ on the three seeds and the implied room does not move
(Table~\ref{tab:app_lsq}). The law over-predicts the LSQ models, and it over-predicts their full-precision siblings and
a fine-tune of the same length just as much, because the MLP family's room (0.17--0.28) sits below the pooled 0.33.
Naive quantization-aware training (seed 0) collapses its own full-precision weights and is not scored. On the A100
platform the LSQ models' full-precision tolerance is 0.233 / 0.238 / 0.214, against 0.137--0.142 for models trained in
full precision. 

\begin{table}[H]
\caption{\textbf{LSQ widens tolerance by lowering the sensitivity; the room stays put.} Sensitivity $S$ and
$S_{\text{mix}}$; implied room $\sigma_{1/2}\cdot S$; pred/meas = the law ($\delta^\star=0.332$, refit on the
15-model set) over measured tolerance. Comparators, both trained in full precision: a fine-tune of the same length
(s0) and the MLP w256 siblings. L40S platform.} 
\label{tab:app_lsq}
\centering\footnotesize
\begin{tabular}{@{}lcccc@{}}
\toprule
Model & $S$ & $S_{\text{mix}}$ & implied room & pred/meas ($S$) \\
\midrule
LSQ-QAT s0 / s1 / s2 & 0.92 / 1.21 / 0.95 & 0.55 / 0.95 / 0.69 & 0.21 / 0.28 / 0.22 & 1.58 / 1.17 / 1.55 \\
fine-tune s0 & 1.57 & 1.09 & 0.21 & 1.57 \\
MLP w256 s0 / s1 / s2 & 1.29--2.05 & 1.05--1.86 & 0.19 / 0.27 / 0.17 & 1.73 / 1.22 / 1.97 \\
\bottomrule
\end{tabular}
\end{table}
\subsection{Distances measured at one noise level do not predict tolerance}\label{app:widthlaw_negative}

Two alternatives to the sensitivity law, both stated before their records, predict tolerance from a distance measured
along the compression direction at a single noise level, and both fail for the same reason. The radius law (P4 in
Table~\ref{tab:app_prereg}) gives $R^2(\log)=-0.84$: at noise 0.1 the MLP models return from the grid cap, yet their
$\sigma_{1/2}$ is among the smallest. The zero-parameter width of explanation (B) is off by $\times8.9$ on the new seeds
(Table~\ref{tab:app_ab}). The settling shift grows faster than linearly with the noise level, so a distance measured at
one level does not transfer to the tolerance. 

\subsection{Fidelity flags collapse with a per-model cut, and the halting head is safe on surviving models}\label{app:widthlaw_detector}

Latent fidelity $\varphi$ separates collapsed from surviving models, pooled over the six puzzle models and MDEQ, with
AUROC 0.98 on the L40S platform (0.97 on the A100 platform; 0.96--1.00 per family), and a cut fitted on the other
families flags collapse in the held-out one with accuracy 0.94 (0.91). The collapse midpoint is model-specific, 0.579
(TRM Sudoku MLP), 0.716 (TRM Sudoku attention), 0.752 (HRM Sudoku), 0.820 (TRM Maze) and 0.897 (HRM Maze), one shared
midpoint is rejected even after dispersion scaling, and within a task the sensitivity does not predict the midpoint (13
Sudoku variants, $R^2$ 0.02). A cut fitted on 50 labelled examples, on one random half of each model's compressed models
and scored on the other half (200 repeats, A100 platform), reaches mean accuracy 1.000 (HRM Sudoku), 0.990 (TRM Sudoku
attention), 0.988 (TRM Maze), 0.972 (HRM Maze), 0.958 (TRM Sudoku MLP) and 0.85--1.00 on the Sudoku variants; it
transfers to a held-out quantizer family when the calibration set reaches near the boundary, and a calibration on noise
levels does not substitute for in-family models. 
The split is not seeded, so reruns of it differ by up to 0.02. 

\paragraph{The halting head as a defer rule.} The head's own rule (sigmoid of the halting logit above 0.5, never
recalibrated) answers wrongly in at most 0.7\% of the cases where it fires and keeps at least 99\% of correct answers,
on all 43 surviving Sudoku models of the three released checkpoints; near the collapse point the risk stays at or below
7.2\% for attention and HRM. On collapsed models it is not a guard, and a threshold recalibrated at full precision to
1\% risk is fragile on surviving models (TRM-attention w3c 11\%). 

\paragraph{The push at the settled state (post hoc).}\label{app:return_posthoc}
After P1 failed we asked what does separate collapsed from surviving models. The push at the full-precision settled
state, $\delta=\lVert f_{\tilde\theta}(z^\star)-f_\theta(z^\star)\rVert/\lVert z^\star\rVert$, separates them with AUROC
0.97 (L40S) and 0.98 (A100), and 0.97--1.00 within each family; its cut is model-specific (0.29--0.80), and sensitivity
times noise level tracks it (Spearman 0.91). The linear-response shift $\delta/(1-g)$, with $g$ the full-precision
loop's gain along the error, does worse (AUROC 0.89), and $g$ alone carries nothing (0.51). Collapse is thus decided at
the settled state by the push against a model-specific cut, which is the room. 
\FloatBarrier

\section{Looped language models}\label{app:lms}
\label{app:lms-tab1}

We quantize the looped block only: Huginn-3.5B \citep{geiping2025recurrentdepth} at 32 recurrences, Ouro-1.4B and
Ouro-2.6B \citep{zhu2025ouro} at their trained 4 loops, and Recurrent-OLMo-2 \citep{mcleish2025retrofitted} at 32, on
GSM8K \citep{cobbe2021gsm8k} with a zero-shot chat prompt (Huginn, Ouro) or 2-shot (Recurrent-OLMo-2), greedy, 512 new
tokens. Formats are w8c, w4g32, w4g128, w4c, w4a, w3a and w4t (Appendix~\ref{app:protocol-quant}). Every accuracy is
paired with a bf16 reference from the same platform, batch and examples. Table~\ref{tab:lms} uses GSM8K test[:100],
batch 16, on the L40S platform; ``+precise'' hands off at loop 16 of 32 (Huginn, Recurrent-OLMo-2) or 2 of 4 (Ouro), and
a model \emph{settles} when the states of its surviving compressed models return under full-precision loops and its
full-precision accuracy does not fall with extra loops. Fidelity orders collapse within every model except Ouro-2.6B,
whose w4a model collapses at a higher fidelity (0.866) than its surviving w4c model (0.852). 

\subsection{The return test does not separate collapse on looped language models}\label{app:lms_return}

Two predictions were written before any return number existed: P1, that one pooled cut on $\rho$ separates collapsed
from surviving models (AUROC and leave-one-model-out $\ge0.95$, and better than $\varphi$), and P2, that per-example
return predicts per-example correctness (pooled AUROC $\ge0.8$). Both fail (Table~\ref{tab:app_lm_ret}): $\rho$ scores
0.700 pooled and 0.708 leave-one-model-out against 0.971 and 0.833 for $\varphi$, and answer-level return predicts
correctness with AUROC 0.616 ($t_c$: 0.573) over the 1062 problems of the 18 models with mixed outcomes. The reason is
the one the puzzle models gave. In Huginn and Recurrent-OLMo-2 the full-precision loops recover the compressed hand-off
state of every model, collapsed ones included (teacher-forced return 85--100\% and 100\%; +precise accuracy 37--46
against bf16 41--42), so a return test cannot tell collapsed from surviving models, since the state stays recoverable
and the compressed loop, run to the end, settles in the wrong place. Ouro does not settle in this sense. Two
full-precision loops return only part of its surviving states (w4g32 73\% and 95\%; w8c 100\%), and $\rho$ orders compressed models
within Ouro-1.4B only. At the answer level, full-precision final loops still repair much of Ouro-1.4B w4c (7 to 66) and
Ouro-2.6B w3a (0 to 50). 

\begin{table}[H]
\caption{\textbf{Return test on all 24 compressed looped language models: in Huginn and Recurrent-OLMo-2, collapsed
states return nearly as often as surviving ones.} Accuracy \%; collapse = retained below 50\% ($\dagger$); +precise =
second half of the loops in bf16; return (teacher-forced) and return (answer) = teacher-forced and answer-level return. L40S
platform, transformers 4.55.4, GSM8K test[:100], batch 16.} 
\label{tab:app_lm_ret}
\centering\scriptsize
\setlength{\tabcolsep}{3.5pt}
\begin{tabular}{@{}llccccccc@{}}
\toprule
Model & Format & bf16 & compressed & +precise & $\varphi$ & $\rho$ & return (teacher-forced) & return (answer) \\
\midrule
Huginn & w8c & 42 & 36 & 41 & 1.000 & 0.25 & 100\% & 80\% \\
Huginn & w4g32 & 42 & 35 & 39 & 0.978 & 0.25 & 97\% & 67\% \\
Huginn & w4c & 42 & 33 & 37 & 0.929 & 0.25 & 91\% & 55\% \\
Huginn & w3a$^\dagger$ & 42 & 3 & 40 & 0.782 & 0.25 & 85\% & 47\% \\
Huginn & w4t$^\dagger$ & 42 & 2 & 38 & 0.696 & 0.25 & 89\% & 54\% \\
\midrule
Recurrent-OLMo-2 & w8c & 41 & 43 & 43 & 1.000 & 0.25 & 100\% & 86\% \\
Recurrent-OLMo-2 & w4g32 & 41 & 42 & 42 & 0.984 & 0.25 & 100\% & 87\% \\
Recurrent-OLMo-2 & w4a & 41 & 34 & 44 & 0.961 & 0.25 & 100\% & 89\% \\
Recurrent-OLMo-2 & w4c & 41 & 35 & 46 & 0.946 & 0.25 & 100\% & 83\% \\
Recurrent-OLMo-2 & w3a$^\dagger$ & 41 & 2 & 45 & 0.790 & 0.25 & 100\% & 85\% \\
Recurrent-OLMo-2 & w4t$^\dagger$ & 41 & 0 & 43 & 0.114 & 0.25 & 100\% & 85\% \\
\midrule
Ouro-1.4B & w8c & 73 & 77 & 79 & 0.997 & 0.25 & 100\% & 86\% \\
Ouro-1.4B & w4g32 & 73 & 75 & 74 & 0.886 & 0.72 & 73\% & 73\% \\
Ouro-1.4B & w4g128 & 73 & 69 & 75 & 0.854 & 0.97 & 52\% & 68\% \\
Ouro-1.4B & w4a & 73 & 55 & 72 & 0.786 & 1.02 & 45\% & 62\% \\
Ouro-1.4B & w4c$^\dagger$ & 73 & 7 & 66 & 0.735 & 1.52 & 22\% & 57\% \\
Ouro-1.4B & w3a$^\dagger$ & 73 & 0 & 0 & 0.181 & 2.29 & 0\% & 0\% \\
Ouro-1.4B & w4t$^\dagger$ & 73 & 0 & 0 & 0.005 & 4.26 & 0\% & 0\% \\
\midrule
Ouro-2.6B & w8c & 81 & 81 & 80 & 0.999 & 0.25 & 100\% & 87\% \\
Ouro-2.6B & w4g32 & 81 & 66 & 73 & 0.923 & 0.46 & 95\% & 74\% \\
Ouro-2.6B & w4a$^\dagger$ & 81 & 34 & 39 & 0.866 & 0.64 & 72\% & 36\% \\
Ouro-2.6B & w4c & 81 & 78 & 74 & 0.852 & 0.79 & 67\% & 73\% \\
Ouro-2.6B & w3a$^\dagger$ & 81 & 0 & 50 & 0.348 & 1.87 & 2\% & 43\% \\
Ouro-2.6B & w4t$^\dagger$ & 81 & 0 & 0 & 0.011 & 2.71 & 0\% & 0\% \\
\bottomrule
\end{tabular}
\end{table}
\subsection{Huginn and Recurrent-OLMo-2 settle; Ouro does not}\label{app:lms_settle}

Ouro is trained at 4 loops and loses accuracy past them even in bf16: Ouro-1.4B peaks early (44.0/79.5/75.5/77.0 at
1/2/3/4 loops, A100 platform, $n{=}200$), Ouro-2.6B at 4/5/6 loops scores 69.0/66.8/64.0 on development rows and
65.4/65.2/62.0 held out, and a sixth loop costs Ouro-1.4B 12.6 (development) and 14.2 (held out) against 4 loops (L40S
platform, $n{=}500$ per split). Running a surviving compressed Ouro with its last loops at INT8 gives no held-out gain
over the compressed model alone at the same weight traffic: Ouro-1.4B w4g128 with INT8 loops 2--3, $+0.2$ ($p{=}1.0$);
Ouro-2.6B w4c $+2.0$ ($p{=}0.41$), w4g128 $-0.4$. On a loop that does not settle nothing guarantees that precise final
loops pull the state back, and surviving models have little gap to recover. Recurrent-OLMo-2, by contrast, keeps its
accuracy and fidelity across loop counts in bf16 and compressed (Table~\ref{tab:app_olmo}), and Huginn's authors report
path independence of its recurrence in full precision \citep{geiping2025recurrentdepth}. 
The fidelity column of Table~\ref{tab:app_olmo} is measured on $n{=}200$ rows and accuracy on $n{=}100$ rows. 

\begin{table}[H]
\caption{\textbf{Recurrent-OLMo-2 keeps its accuracy and fidelity across loop counts.} Accuracy \% with $\varphi$ in
parentheses; teacher-forced KL(bf16$\|$q): w4g32 0.02, w4c 0.09, w4t 9.6. A100 platform, GSM8K test[:100], 2-shot
greedy; Wilson half-width $\approx\pm10$ points.} 
\label{tab:app_olmo}
\centering\footnotesize
\begin{tabular}{@{}lccccc@{}}
\toprule
Loops & bf16 & w8c & w4g32 & w4c & w4t \\
\midrule
4 & 44.0 & 43.0 (1.00) & 41.0 (0.98) & 31.0 (0.942) & 0.0 (0.12) \\
8 & 44.0 & 51.0 (1.00) & 40.0 (0.98) & 37.0 (0.947) & 0.0 (0.11) \\
16 & 45.0 & 47.0 (1.00) & 41.0 (0.98) & 36.0 (0.947) & 0.0 (0.12) \\
32 & 42.0 & 44.0 (1.00) & 42.0 (0.98) & 38.0 (0.946) & 0.0 (0.11) \\
\bottomrule
\end{tabular}
\end{table}
\subsection{Fidelity orders collapse within each looped language model, with a cut per model}\label{app:lms_detect}

Within each model, fidelity orders compressed models without error on the deduplicated multi-grid pool (112 compressed
models, 77 surviving, 35 collapsed; AUROC 1.000 per model), but each model's boundary is only bracketed, Ouro-1.4B
[0.737, 0.885], Ouro-2.6B [0.427, 0.808], Recurrent-OLMo-2 [0.791, 0.942] and Huginn [0.784, 0.897]; the brackets
intersect near 0.8 because they are wide, and neither answer length, sensitivity nor answer-logit margin predicts a
model's midpoint. With 50 labelled examples on disjoint compressed models, a per-model cut reaches mean accuracy 0.999
(Ouro-1.4B), 0.996 (Ouro-2.6B), 0.982 (Huginn) and 0.889 (Recurrent-OLMo-2, with one collapsed model near the
boundary). 
This pool mixes three platforms within a model (the two of Appendix~\ref{app:protocol-stats} and an earlier one with
H100 and A100-80GB cards); restricted to one platform per model it keeps AUROC 1.000 per model, with the brackets not
re-checked, and its calibration split is unseeded like that of Appendix~\ref{app:widthlaw_detector}. 

\paragraph{KL from bf16.} On the saved teacher-forced records of one platform (A100; 39 compressed models of the four
looped language models, 14 collapsed, each paired with a bf16 reference at the same batch), $\mathrm{KL}(\text{bf16}\,\|\,q)$
separates collapsed from surviving models with pooled AUROC 1.000 and a gap (collapsed $\ge0.214$, surviving
$\le0.174$), leave-one-model-out accuracy 0.949; $1-\varphi$ gives 0.966 / 0.846 and relative $\Delta$NLL 1.000 / 1.000
on the same models. 

\paragraph{Loss and collapse.} A small loss increase coexists with collapsed generation. For Recurrent-OLMo-2 w3a at
32 loops (A100 platform, $n{=}200$, batch 8 for both), the teacher-forced gold NLL rises by 0.805 nats, top-1 agreement
with bf16 stays at 0.709 and likelihood-scored answer accuracy is 83.0, yet greedy accuracy is 2.0 against 42.5 for
bf16. On a separate Huginn/Ouro grid, collapsed models start at a relative NLL increase of $+16\%$ (Huginn w3a, keeping
15\%) and surviving models reach $+20\%$ (Ouro-1.4B w8t at 1 loop, keeping 59\%), so no absolute loss threshold
separates them. 
That grid ran on a different transformers version; $n{=}50$--200 per model. 
\FloatBarrier

\section{Controller details}\label{app:controller}

Throughout, a \emph{step} is one supervision step of TRM or HRM, the controller's loop.

\subsection{The protocol of Table~\ref{tab:controller}, and the rows it leaves out}\label{app:controller-tab2}

The models are TRM Sudoku-MLP, TRM Sudoku-attention and HRM Sudoku, as released checkpoints and as six trained seeds
(attn512 and mlp512, s0--s2), which gives the 20 quantized Sudoku-Extreme models, at a budget of 64 cost units. Every configuration
is chosen on rows 0--999 and scored on rows 1000--1999 ($n{=}1000$ per model, paired), on the A100 platform; brackets
are paired puzzle-bootstrap 95\% intervals, except the random-gate range (2.5--97.5 percentile of 1000 random sets).
Two rows are left out of Table~\ref{tab:controller}: stopping and sampling without finishing reaches 67.3 at cost 40.3,
and the sampling add-on with full-precision finishing reaches 78.7 at cost 39.6, against 78.7 at cost 28.4 with 8-bit
finishing (8-bit minus full precision $-0.00$ [$-0.22$, $+0.22$]). Among stopping rules, only the halting head improves
on the 16-step protocol depth (Table~\ref{tab:app-stoprules}); the frozen step-size threshold stops earlier at a small
loss, and waiting for the step size to stop shrinking loses eleven points. 

\begin{table}[h]
\centering
\caption{\textbf{Only the halting head stops early and gains accuracy over the protocol depth.} Change in held-out
exact accuracy (points) against running every puzzle to the 16-step protocol depth, and mean steps per puzzle; means
over the 46 models of the grid (compressed, full-precision and weight-noise models). Every rule is capped at 16 steps
and frozen on full-precision development rows unless marked. Held-out rows 1000--1999 for Sudoku ($n{=}1000$) and
500--999 for Maze-Hard ($n{=}500$); one platform (A100 platform).}
\label{tab:app-stoprules}
\small
\begin{tabular}{@{}lcc@{}}
\toprule
Stopping rule & $\Delta$ accuracy & Mean steps \\
\midrule
Halting head, rule frozen per model                        & $+1.23$  & 8.15 \\ 
Step-size threshold, frozen on full precision              & $-0.36$  & 6.04 \\ 
Step-size threshold, tuned per model                       & $-0.43$  & --   \\ 
Stop when the step size stops shrinking                    & $-11.2$  & 7.2  \\ 
\bottomrule
\end{tabular}
\end{table}
\FloatBarrier
\subsection{The halting head stops early and keeps answers that later loops lose}\label{app:controller-head}

TRM's halting head is one linear layer on the first position of the high-level state, $q = W z_H[0] + b$, with a
halting and a continue logit \citep{jolicoeur2025trm}; HRM's is the same readout trained by its ACT objective
\citep{wang2025hrm}. The quantizers compress it with every other linear layer. One firing rule per model is chosen on
the full-precision model's development rows, as the rule with the fewest mean steps within 0.5 points of the 16-step
protocol, from three families (native: halting logit $>0$, or halting $>$ continue for HRM; patience $p$: the native
condition at $p$ consecutive steps; a minimum-step guard), and then frozen for every compressed version: patience 2
for TRM Sudoku-MLP, native for TRM Sudoku-attention, HRM's own ACT rule, and a minimum of 2 steps for the mazes.
On surviving models stopping saves steps at almost no cost: TRM Sudoku stops 1.75--2.2$\times$ earlier at a loss of at
most 0.9 points, HRM Sudoku's rule is lossless at 1.5$\times$, and the mazes stop 4.9--7.9$\times$ earlier, which no
fixed shorter depth matches (a fixed depth equal to the head's mean steps loses 4--5 points on TRM and 2.4 on HRM).
Past the room, stopping raises accuracy, because trajectories reach the answer and then settle off target: TRM
Sudoku-MLP with weight noise 0.15 goes from 29.8 to 52.9, MLP w3g32 from 51.2 to 58.4, and HRM Maze with activation
noise 1.0 from 40.1 to 55.4 (L40S platform, first 1000 test rows). On held-out rows of the A100 grid the frozen rule is
non-inferior to the 16-step protocol in 40 of 46 models, better in 12 (up to $+17.1$, on models that overthink) and
worse in 4 ($-0.4$ to $-1.4$). A head silenced by its own low-bit weights (HRM Maze w3t, w2a, w2g32) is repaired by
keeping its 1{,}026 parameters at full precision, a fix Table~\ref{tab:controller} does not use. A threshold on the raw
halting logit lifts the released MLP further past the room than the patience rule ($+12.9$ [$+10.8$, $+15.0$] and
$+12.3$ on weight noise 0.15, s0/s2) at a cost of 1.0 point at full precision on surviving models.
Without a head, the step-size rule stops at the first step with $\|z_H^{(k)}-z_H^{(k-1)}\|/\|z_H^{(k)}\|<\epsilon$,
capped at 16. With $\epsilon$ frozen on full-precision development rows it scores $-0.36$ at 6.04 mean steps against
$+1.23$ at 8.15 for the head; it is worse than the head by interval in 22 of 46 models and non-inferior in 17, tuning
$\epsilon$ per model does not help, and it gives up most of the head's past-room gains (attn512 s2: $+1.2$/$-2.0$
against $+7.1$/$+6.7$), and matches or beats the head on MLP with weight noise 0.15 ($+8.3$/$+8.8$ against
$+6.2$/$+8.2$). 

\FloatBarrier
\subsection{Finishing schedules and the cost of a finishing step}\label{app:controller-finishing}
\label{app:controller-cost}

Finishing records store correctness after $j\le4$ finishing steps started from the compressed state after every
compressed step $t\le16$. After a stop the controller may finish with $j_E\in\{0,1,2,4\}$ steps; at the cap it finishes
with $j_F\in\{0,1,2,4\}$ steps, or $j_F\in\{0,1,2\}$ with the sampling add-on, whose rollout end states have records for
$j\le2$ and for full precision only, which can only lower the 8-bit add-on row. Finishing every puzzle is $t$
compressed steps followed by $j$ finishing steps. On TRM Sudoku-attention the frozen development pick, 224 w3c steps
followed by 32 w8c steps, scores 91.9 on held-out rows against 87.2 for w3c alone at matched weight traffic (299 steps
as counted in that experiment; the cost model below would charge 309), a gain of $+4.7$ (50/3 discordant puzzles, exact
McNemar $p{=}5.5\times10^{-12}$), and beats full precision at its best depth up to 512 by $+3.8$ ($p{=}2.5\times10^{-5}$)
(L40S platform). Spending weight traffic on 8-bit finishing steps thus beats spending it on more compressed steps. A
$b$-bit format with group size $g$ stores $b_{\text{eff}} = b + 16/g$ bits per weight, and a finishing step at $b_f$ bits
costs $r = b_f/b_{\text{eff}}$ compressed steps (Table~\ref{tab:app-ratio}). A w3c puzzle whose head fires at step 9 and
then takes two finishing steps costs $9 + 2\times2.67 = 14.3$ units with 8-bit finishing and $9 + 2\times10.7 = 30.3$
with full precision. Weight-noise and full-precision models carry full-size weights ($r{=}1$) and are left out of the
cost headline; activation traffic, kernel launches and switching weight sets are not charged.

\begin{table}[h]
\centering
\caption{\textbf{An 8-bit finishing step costs two to four compressed steps; a full-precision one costs eight to
sixteen.} Cost $r$ of one finishing step in compressed steps, $r=b_f/b_{\text{eff}}$, by compressed format;
$b_{\text{eff}}$ counts the bits read per weight including group scales. Weight traffic only.}
\label{tab:app-ratio}
\small
\begin{tabular}{@{}lccccc@{}}
\toprule
Compressed format & w4 & w3 & w3g32 & w2g32 & w2 \\
\midrule
$b_{\text{eff}}$ & 4 & 3 & 3.5 & 2.5 & 2 \\
Full-precision finishing, $r=32/b_{\text{eff}}$ & 8 & 10.7 & 9.1 & 12.8 & 16 \\
8-bit finishing, $r=8/b_{\text{eff}}$ & 2 & 2.67 & 2.29 & 3.2 & 4 \\
\bottomrule
\end{tabular}
\end{table}
\FloatBarrier
\subsection{The controller's gain comes from collapsed models}\label{app:controller-cells}

Table~\ref{tab:app-cells} lists every quantized model of the A100 grid at a budget of 64 with full-precision finishing.
Over its 24 models the means are fixed depth 62.7 at cost 47.3; stopping 64.4 at 20.3; finishing every puzzle 73.4 at
43.9 (3.58 finishing steps per puzzle); the controller 76.1 at 24.9 (1.23); the low-confidence selector 72.6 at 26.2;
stopping and sampling 68.0 at 36.3; the controller with the sampling add-on 78.7 at 36.4 (at a budget of 24: fixed
depth 60.5, controller 72.2 at 16.7, add-on 72.8 at 17.6). Table~\ref{tab:controller} uses the 20 Sudoku models, the
ones with 8-bit finishing records. Per model, the controller with 8-bit finishing is above fixed depth with an interval
above zero in 10 of the 20, from $+0.8$ (attn512 s1 w3c) to $+63.7$ (mlp512 s2 w3c); its interval includes zero in 6
and lies below zero in 4 (TRM Sudoku-attention w3c $-2.4$, mlp512 s1 w4c $-0.6$, attn512 s0 w4c and mlp512 s2 w4c
$-0.4$). The large gains sit on the collapsed models, where the development rows pick finishing, and on attn512 s2,
where stopping alone carries them; surviving models stay within 2.4 points of fixed depth at a fraction of its cost.
With the sampling add-on every one of the 27 Sudoku models (weight-noise models included) ends above fixed depth with
an interval above zero, from $+1.8$ to $+63.7$; under unit costing over all 46 models the add-on reaches 77.4 against
64.7 for fixed depth and is at or above it in every model at budgets 16, 64 and 256.
The four Maze models also have 8-bit finishing records. Over them, at a budget of 32 and with the same selection rule,
fixed depth scores 70.9 at a cost of 9.0, and the controller with 8-bit finishing scores 78.5 at a cost of 9.1, a gain
of 7.6 points [6.2, 9.0]; full-precision finishing reaches 78.3 at a cost of 20.0, so the 8-bit copy matches it at
under half the cost.

\begin{table}[h]
\centering
\caption{\textbf{The controller's gain sits on the collapsed models; surviving models keep their accuracy at a
fraction of the cost.} Each entry is held-out accuracy (\%) / mean cost in compressed steps, at a budget of 64, with
full-precision finishing throughout; $r$ is the cost of one full-precision finishing step. Stop: stop at the halting
head. Finish all: finish every puzzle. Controller: stop, then finish. Low confidence: finish when the head's confidence
is low. Stop, sample: stop and sample, no finishing. +Sampling: the controller with the sampling add-on.
Configurations are chosen on development rows (Sudoku 0--999, mazes 0--499) and scored on held-out rows (Sudoku
1000--1999, mazes 500--999). One platform (A100 platform, A100-40GB cards).}
\label{tab:app-cells}
\scriptsize
\setlength{\tabcolsep}{3pt}
\begin{tabular}{@{}llrccccccc@{}}
\toprule
Model & Format & $r$ & Fixed & Stop & Finish all & Controller & Low confidence & Stop, sample & +Sampling \\
\midrule
TRM-MLP & w3g32 & 9.1 & 59.1/32 & 61.5/22 & 74.2/51 & 74.2/51 & 74.1/32 & 71.2/52 & 76.4/56 \\
TRM-MLP & w4c & 8.0 & 77.5/64 & 77.5/19 & 76.5/48 & 77.5/19 & 76.5/18 & 82.6/36 & 82.6/36 \\
TRM-MLP & w4t & 8.0 & 7.0/64 & 7.0/64 & 67.3/42 & 67.3/42 & 66.8/40 & 7.7/64 & 67.3/42 \\
TRM-attn & w3c & 10.7 & 82.9/64 & 80.5/17 & 80.6/58 & 80.5/17 & 79.7/40 & 86.3/32 & 86.9/34 \\
TRM-attn & w4c & 8.0 & 83.8/64 & 83.7/15 & 79.4/47 & 83.7/15 & 79.8/15 & 86.8/32 & 86.8/32 \\
HRM & w2g32 & 12.8 & 20.8/64 & 20.5/62 & 37.0/52 & 37.0/52 & 34.6/55 & 26.0/64 & 37.0/52 \\
HRM & w4c & 8.0 & 64.9/64 & 64.9/27 & 60.2/47 & 64.9/27 & 60.5/26 & 67.0/49 & 67.0/49 \\
HRM & w4t & 8.0 & 64.0/64 & 64.0/28 & 60.5/48 & 64.0/28 & 60.5/26 & 66.5/50 & 66.5/50 \\
attn512 s0 & w4c & 8.0 & 84.0/64 & 83.6/16 & 78.3/47 & 83.6/16 & 78.2/16 & 87.4/33 & 87.3/33 \\
attn512 s0 & w3c & 10.7 & 82.0/64 & 81.9/18 & 77.5/59 & 81.9/18 & 77.5/22 & 88.5/45 & 88.5/45 \\
attn512 s1 & w4c & 8.0 & 79.7/64 & 81.2/17 & 74.5/46 & 81.2/17 & 76.7/17 & 85.4/37 & 85.4/37 \\
attn512 s1 & w3c & 10.7 & 74.1/64 & 74.9/20 & 73.3/57 & 74.9/20 & 75.0/26 & 79.7/44 & 79.7/44 \\
attn512 s2 & w4c & 8.0 & 64.7/32 & 77.7/19 & 66.4/45 & 77.7/19 & 73.5/16 & 83.1/40 & 83.1/40 \\
attn512 s2 & w3c & 10.7 & 50.2/12 & 75.7/22 & 65.8/50 & 75.7/22 & 72.5/27 & 80.8/44 & 80.8/46 \\
mlp512 s0 & w4c & 8.0 & 96.7/64 & 96.7/7 & 92.5/48 & 96.7/7 & 92.5/9 & 99.0/11 & 99.0/11 \\
mlp512 s0 & w3c & 10.7 & 14.2/48 & 14.2/48 & 67.1/44 & 67.1/44 & 57.0/55 & 21.3/64 & 67.1/44 \\
mlp512 s1 & w4c & 8.0 & 93.8/64 & 93.2/9 & 88.6/44 & 93.2/9 & 89.6/9 & 97.3/15 & 98.1/42 \\
mlp512 s1 & w3c & 10.7 & 20.6/64 & 21.1/16 & 69.2/44 & 69.2/44 & 64.0/46 & 25.9/17 & 69.2/44 \\
mlp512 s2 & w4c & 8.0 & 96.6/64 & 96.2/8 & 91.5/48 & 96.2/8 & 91.5/9 & 98.4/12 & 98.4/12 \\
mlp512 s2 & w3c & 10.7 & 4.9/16 & 4.9/16 & 67.7/44 & 67.7/44 & 50.3/47 & 5.6/64 & 67.7/44 \\
TRM Maze & w2a & 16.0 & 68.2/12 & 68.4/5 & 83.2/17 & 83.2/17 & 83.0/36 & 68.8/31 & 83.2/17 \\
TRM Maze & w4t & 8.0 & 85.0/12 & 85.0/2 & 85.2/9 & 85.2/9 & 85.0/2 & 85.0/2 & 85.2/9 \\
HRM Maze & w2a & 16.0 & 69.4/8 & 69.4/7 & 72.8/34 & 72.8/31 & 71.8/8 & 68.8/8 & 72.8/31 \\
HRM Maze & w2g32 & 12.8 & 61.0/4 & 61.0/3 & 72.2/28 & 72.2/23 & 72.4/30 & 62.0/25 & 72.2/23 \\
\bottomrule
\end{tabular}
\end{table}
\FloatBarrier
\subsection{The halting head picks the puzzles to finish better than chance}\label{app:controller-random}
\label{app:controller-selectors}

A threshold on the head's confidence at the stop step (finish the puzzle with $j$ full-precision steps when the halting
logit, or halting minus continue for HRM, is below $\theta$, with $D_0$, $\theta$ and $j$ chosen per budget on
development rows) beats fixed depth by $+9.9$ [$+9.5$, $+10.4$] over the 24 models but trails the controller by 3.5
points at similar cost and the add-on by $6.1$ [$5.6$, $6.5$]. The controller wins because its selection falls back
to finishing everything where a model has collapsed and to stopping alone where it survives. A configuration that
finishes some puzzles and not others is picked in only 2 of 24 models. The head's choice of puzzles within a model
therefore needs its own control, which holds the number of finished puzzles and the cost fixed and changes only which
puzzles are finished. The \emph{gated} family finishes exactly the puzzles on which the head has not fired by a cap
$D_0\le16$, with $j\in\{1,2,4\}$ finishing steps chosen per model on development rows and scored on held-out rows; the
random arm keeps each puzzle's stop step and permutes the per-puzzle finishing-step counts across puzzles, 1000
permutations per model. With full-precision finishing over the 24 models the head gate scores 73.9 against 72.0
[71.9, 72.1] at cost 23.5 (head minus random $+1.88$ [$+1.75$, $+2.00$]); the gate is active in 17 models, and the
head is ahead of the random mean in all 17 and above its 97.5th percentile in 16 (HRM Maze w2a sits on the boundary,
$p\approx0.03$). With 8-bit finishing over the 20 Sudoku models the head scores 73.2 against 71.3 [71.2, 71.4] at cost
10.6, above the random range in all 15 active models ($+0.7$ to $+5.4$); at a budget of 24 the results agree ($+1.48$
and $+1.90$). The same finishing steps placed at random buy about two points less, so the head's firing carries
information about which puzzles still need finishing (Table~\ref{tab:app-random}).

\begin{table}[h]
\centering
\caption{\textbf{The head chooses which puzzles to finish better than chance in every model where the gate is
active.} Gated finishing (finish the puzzles on which the head has not fired by $D_0$) against random sets of the same
size at identical cost; held-out accuracy (\%), budget 64. Random: mean [2.5, 97.5 percentile] over 1000
permutations; ``Finished'' is the fraction of puzzles finished. Rows at 1.00 finish every puzzle, where the two
arms coincide. One platform (A100 platform, A100-40GB cards).}
\label{tab:app-random}
\scriptsize
\setlength{\tabcolsep}{3pt}
\begin{tabular}{@{}llcccccc@{}}
\toprule
 & & \multicolumn{3}{c}{Full-precision finishing} & \multicolumn{3}{c}{8-bit finishing} \\
\cmidrule(lr){3-5}\cmidrule(l){6-8}
Model & Format & Head & Random & Finished & Head & Random & Finished \\
\midrule
TRM-MLP & w3g32 & 74.1 & 69.2 [68.2, 70.1] & 0.64 & 74.8 & 69.4 [68.3, 70.4] & 0.64 \\
TRM-MLP & w4c & 76.5 & 74.1 [73.6, 74.6] & 0.27 & 76.8 & 74.2 [73.6, 74.7] & 0.27 \\
TRM-MLP & w4t & 66.8 & 66.8 [66.8, 66.8] & 1.00 & 67.6 & 67.6 [67.6, 67.6] & 1.00 \\
TRM-attn & w3c & 76.9 & 72.4 [71.6, 73.3] & 0.27 & 76.9 & 72.5 [71.6, 73.3] & 0.27 \\
TRM-attn & w4c & 79.4 & 78.7 [78.4, 79.0] & 0.21 & 79.6 & 78.7 [78.5, 79.1] & 0.21 \\
HRM & w2g32 & 37.0 & 37.0 [37.0, 37.0] & 1.00 & 36.8 & 36.8 [36.8, 36.8] & 1.00 \\
HRM & w4c & 60.5 & 59.8 [59.5, 60.1] & 0.41 & 60.5 & 59.8 [59.5, 60.1] & 0.41 \\
HRM & w4t & 60.5 & 59.3 [58.8, 59.7] & 0.42 & 60.5 & 59.3 [58.8, 59.7] & 0.42 \\
attn512 s0 & w4c & 77.9 & 76.8 [76.5, 77.2] & 0.23 & 77.8 & 76.8 [76.5, 77.1] & 0.23 \\
attn512 s0 & w3c & 77.4 & 74.3 [73.8, 74.8] & 0.27 & 77.7 & 74.4 [73.8, 75.0] & 0.27 \\
attn512 s1 & w4c & 76.7 & 75.0 [74.6, 75.4] & 0.25 & 76.5 & 74.9 [74.5, 75.3] & 0.25 \\
attn512 s1 & w3c & 74.8 & 71.6 [70.9, 72.2] & 0.29 & 75.1 & 71.7 [71.0, 72.3] & 0.29 \\
attn512 s2 & w4c & 73.5 & 69.9 [69.0, 70.7] & 0.28 & 73.7 & 69.8 [68.9, 70.8] & 0.28 \\
attn512 s2 & w3c & 72.4 & 67.7 [66.9, 68.6] & 0.33 & 72.3 & 67.7 [66.9, 68.6] & 0.33 \\
mlp512 s0 & w4c & 92.5 & 90.2 [89.9, 90.5] & 0.10 & 91.7 & 90.1 [89.9, 90.4] & 0.10 \\
mlp512 s0 & w3c & 67.1 & 67.1 [67.1, 67.1] & 1.00 & 68.7 & 68.7 [68.7, 68.7] & 1.00 \\
mlp512 s1 & w4c & 88.2 & 85.0 [84.6, 85.5] & 0.15 & 89.0 & 87.4 [87.1, 87.7] & 0.12 \\
mlp512 s1 & w3c & 69.2 & 69.2 [69.2, 69.2] & 1.00 & 68.4 & 68.4 [68.4, 68.4] & 1.00 \\
mlp512 s2 & w4c & 91.1 & 89.3 [89.0, 89.6] & 0.11 & 91.0 & 89.3 [89.0, 89.6] & 0.11 \\
mlp512 s2 & w3c & 67.7 & 67.7 [67.7, 67.7] & 1.00 & 68.6 & 68.6 [68.6, 68.6] & 1.00 \\
TRM Maze & w2a & 83.2 & 83.2 [83.2, 83.2] & 1.00 & -- & -- & -- \\
TRM Maze & w4t & 85.2 & 85.2 [85.2, 85.2] & 1.00 & -- & -- & -- \\
HRM Maze & w2a & 72.8 & 71.8 [70.8, 72.6] & 0.90 & -- & -- & -- \\
HRM Maze & w2g32 & 72.2 & 67.3 [65.8, 68.8] & 0.83 & -- & -- & -- \\
\midrule
Mean & & 73.9 & 72.0 [71.9, 72.1] & & 73.2 & 71.3 [71.2, 71.4] & \\
\bottomrule
\end{tabular}
\end{table}
\FloatBarrier
\subsection{Redrawing the rounding beats latent sampling, and the rounding grid itself does not matter}\label{app:controller-sampling}

The add-on row of Table~\ref{tab:controller} uses PTRM-style latent perturbations \citep{sghaier2026ptrm}: $z_L
\mathrel{+}= \mathcal{N}(0,\sigma^2 I)$ before every step with $\sigma{=}0.3$, in rollouts of 16 steps each stopped by
the same head, at most 16 rollouts, tried in order until one fires; if none fires the controller keeps the candidate
with the highest final halting logit. A second way to sample redraws the rounding itself. Stochastic-rounding best-of-16,
with the draw the head trusts most kept, beats the full-precision model sampled with latent noise tuned per model on
rows 0--999 by 5.7 to 7.1 points on held-out rows, and Gaussian weight noise of the rounding draws' variance does as
well as the draws (Table~\ref{tab:app-sampling}). Sampling is an ensemble over weight perturbations with the head as
its verifier \citep{gal2016dropout,maddox2019swag,lakshminarayanan2017ensembles,cobbe2021gsm8k}. Where its draws are
weak it loses (MLP w3g32 $-0.9$ to $-1.3$; w3a $-7$ to $-8$; HRM w2g32 $-3$ to $-5$), and a label-free rule flags those
models. The head's acceptance rate of rounding draws over that of latent draws is at most 0.469 in every losing model
and at least 0.871 in every winning one, and a 0.8 threshold fixed on rows 0--999 classifies all 11 held-out models.
The head is a stronger selector than agreement across draws (AUROC 0.99--1.00 against 0.94--0.98), a majority vote
loses 3--8 points to it, and an 8-bit master copy gives the same accuracy as a full-precision one on development rows,
so the draws can come from the 8-bit copy that finishing already needs. 

\begin{table}[h]
\centering
\caption{\textbf{Redrawing the rounding and keeping the draw the head trusts most beats full precision sampled with
tuned latent noise, and matched Gaussian weight noise does as well as the rounding draws.} Exact accuracy (\%).
Top: held-out rows 1000--1999, $K{=}16$ draws, $\sigma$ tuned on rows 0--999, with paired bootstrap 95\% intervals
for the difference. Bottom: development rows 0--999 only. One platform (L40S platform), $n{=}1000$.}
\label{tab:app-sampling}
\scriptsize
\setlength{\tabcolsep}{3pt}
\begin{tabular}{@{}lccc@{}}
\toprule
 & TRM-MLP w4c & TRM-attn w3c & HRM w3c \\
\midrule
Stochastic rounding, best of 16 (held out)       & 93.7 & 94.2 & 79.2 \\ 
Full precision + latent noise, best-of-16        & 88.0 & 87.1 & 73.3 \\ 
Difference [95\% CI]                             & $+5.7$ [$+4.1$, $+7.4$] & $+7.1$ [$+5.3$, $+9.0$] & $+5.9$ [$+3.8$, $+8.0$] \\ 
\midrule
Stochastic-rounding draws (rows 0--999)          & 86.9 & 86.4 & 71.9 \\ 
Gaussian noise at the draws' variance (rows 0--999) & 87.0 & 85.7 & 71.7 \\ 
\bottomrule
\end{tabular}
\end{table}
\FloatBarrier

\end{document}